\documentclass[10pt,twocolumn,letterpaper]{article}

\usepackage{cvpr}              % To produce the CAMERA-READY version
\usepackage{amsmath,amssymb}
\usepackage{algorithm}
\usepackage{algpseudocode}
\usepackage{fvextra}
\usepackage{xcolor}
\algrenewcommand{\algorithmicrequire}{\textbf{Input:}}
\algrenewcommand{\algorithmicensure}{\textbf{Output:}}
\usepackage{pifont}
\usepackage{url}
\usepackage{multirow}
\usepackage{xurl}
\definecolor{forestgreen2}{rgb}{0.0, 0.61, 0.0}

\definecolor{cvprblue}{rgb}{0.21,0.49,0.74}
\usepackage[pagebackref,breaklinks,colorlinks,allcolors=cvprblue]{hyperref}

\def\confName{CVPR}
\def\confYear{2027}

\title{ARCH3D: Agentic Reconstruction of hierarCHical 3D building models}

\author{
Ruiqi Cui\textsuperscript{*}\qquad
Thor Vestergaard Christiansen\textsuperscript{*}\qquad
Morten R. Hannemose\\
Jan Karlshøj\\
J. Andreas Bærentzen\\
Technical University of Denmark\\
Anker Engelunds Vej 101, 2800 Kongens Lyngby, Denmark\\
{\tt\small \{ruicu,tdvc,mohan,jkar,janba\}@dtu.dk}\\
{\small \textsuperscript{*}Equal contribution.}
}

\begin{document}
\twocolumn[{%
\renewcommand\twocolumn[1][]{#1}%
\maketitle
\vspace{-2.2em}
\begin{center}
  \includegraphics[width=0.8\textwidth]{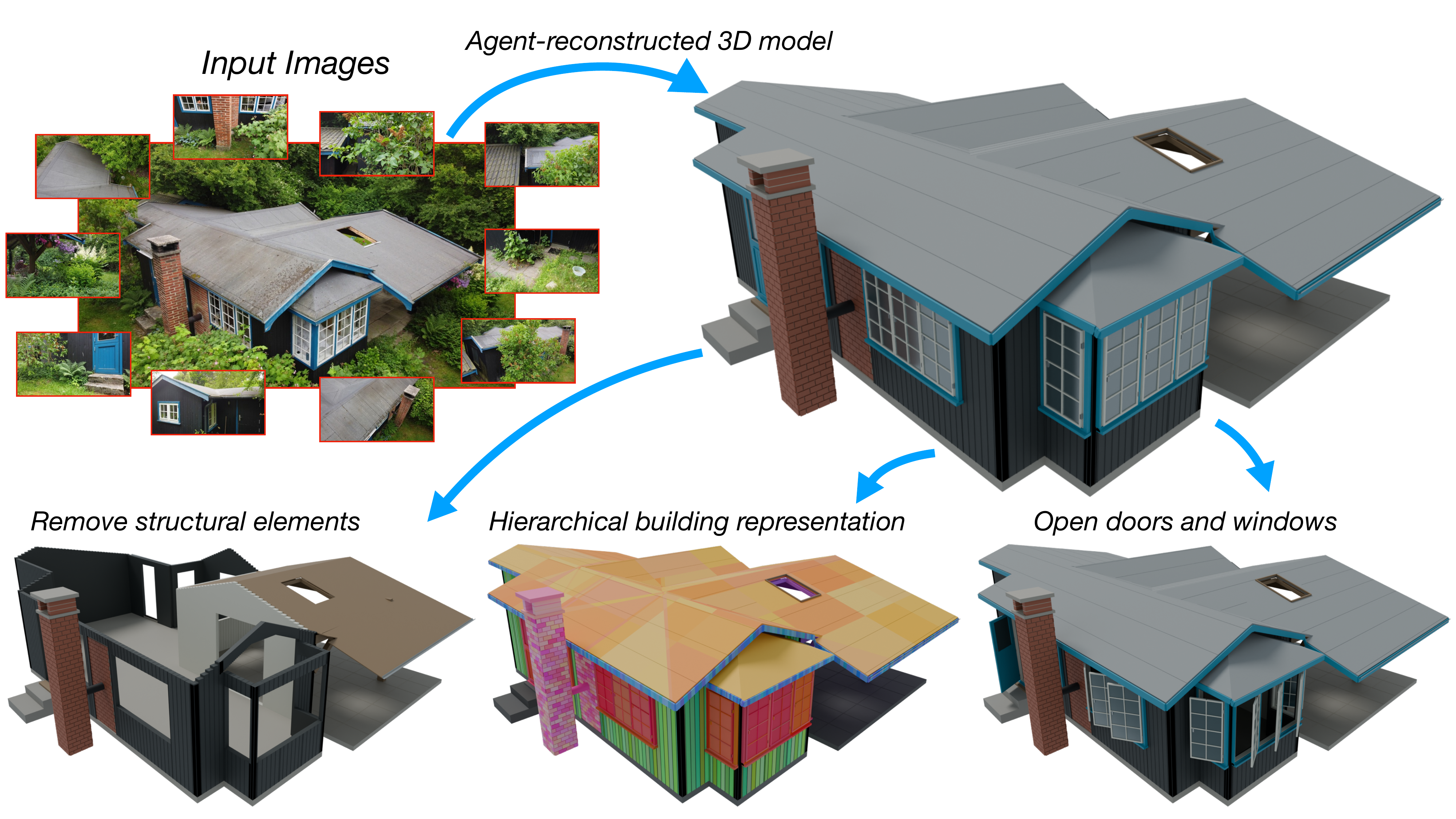}
  \captionof{figure}{Our framework, ARCH3D, generates a hierarchical geometric 3D building model using an AI agent, which takes a set of photographs as input (here, one from the \textit{Cottages} dataset). The 3D building model supports editing and animation of individual structural elements.}
  \label{fig:teaser}
\end{center}
%\vspace{0.6em}
}]
\begin{abstract}
We introduce ARCH3D, a new method for generating hierarchical geometric 3D models of buildings from photographs using an AI agent. Our work is motivated by the increasing focus on recycling materials and repurposing rather than demolishing buildings. Both recycling and repurposing require semantically rich and geometrically accurate building models to assess the types and quantities of structural elements. Purely visual representations, such as those produced by Neural Radiance Fields (NeRFs) or 3D Gaussian Splatting, and 3D surface reconstructions, such as polygonal meshes produced by methods like NeuS, do not inherently provide this information. Our method uses the agent’s ability to interpret photographs and recognize individual objects. The agent also measures building elements using semantic cues from segmentation masks and geometric information from dense COLMAP reconstructions. These measurements and visual observations guide the construction and iterative refinement of the 3D building model.
%Our agentic framework consists of an iterative process, where the agent uses the visual information provided by the photographs and combines it with semantic cues and precise geometric measurements extracted from dense reconstructions and segmentation masks, to construct a hierarchical 3D building model.
%Our agentic workflow consists of an iterative process, where the input photographs are leveraged to construct a hierarchical 3D building model by combining the visual information with additional modalities that provide semantic cues and precise geometric measurements extracted from dense reconstructions and segmentation masks. 
%The availability of camera information further allows us to introduce a refinement loop following the initial reconstruction. 
The hierarchical structure of the 3D building produced by our method makes it straightforward to remove or modify attached objects, temporary installations, and interior furnishings. Furthermore, they contain functional elements such as doors and windows. We evaluate our method on diverse building datasets and compare it with existing work. Our ablation study examines how harness control, GPT model variants, and reasoning effort affect reconstruction quality, alongside their time and token costs.
\vspace{-5mm}

%The models produced by our method go beyond visualisation and contain functional elements such as doors and windows. Their hierarchical structure also makes it straightforward to remove or modify attached objects, temporary installations, and interior furnishings. Furthermore, in an ablation study, we compare a range of VLMs and variations of the agentic workflow.

%The input to our method consists of photographs. Our agentic workflow leverages these inputs to construct a hierarchical 3D building model. Recent Vision-Language Models (VLMs) are capable of extracting salient information about building structure. By combining this information with additional modalities that provide semantic cues and precise geometric measurements, the agent produces detailed hierarchical models with meaningful architectural structure. The availability of camera information further allows us to introduce a refinement loop following the initial reconstruction. In an ablation study, we compare a range of VLMs and variations of the agentic workflow. We then evaluate the strongest configuration on a diverse collection of building datasets and compare it to existing work. The models produced by our method go beyond visualisation and contain functional elements such as doors and windows. Their hierarchical structure also makes it straightforward to remove or modify attached objects, temporary installations, and interior furnishings.
\end{abstract}    
\section{Introduction}
\label{sec:intro}
Recent advances in large vision-language models (VLMs) have enabled 3D model generation through scripts derived primarily from image inputs. Exploiting this new possibility, we propose a novel method for generating geometrically accurate, \textit{hierarchical}, and semantically rich 3D building models called ARCH3D. We consider a model to be hierarchical and semantically rich if it is organized as a tree, or, more generally, a directed acyclic graph, whose leaf nodes contain the geometric primitives and whose intermediate nodes carry semantic meaning. For instance, we might have an intermediate node labelled ``door" whose child nodes would be the geometry that actually represents the door, like ``glass" and ``hinge". And the door node would itself be a child node of a higher-level node such as, e.g., ``openings" (see \cref{fig:pipeline}). %such as e.g. ``south-facing wall''.
The combination of a hierarchy and semantics has many advantages. The semantic labels allow us to locate similar types of nodes with ease: e.g. all windows or power plugs, and the hierarchy ensures that we can easily obtain all the geometry that belongs to the located objects -- even if that geometry may be much further down the hierarchy. This has important downstream applications such as replacing repeated geometry with instances of a single object or removing a group of objects (e.g. all furniture in a given room) simply by hiding a common parent node along with its children, as illustrated in \cref{fig:teaser}.

Hierarchical, semantically rich 3D models provide a foundation for Building Information Modelling (BIM) and the sustainable management of existing buildings. Construction and demolition activities account for more than $35\%$ of waste generated in the EU \cite{european2024addressing}, motivating strategies that extend building lifetimes and recover reusable components. Such strategies require knowledge of the types, quantities, locations, and conditions of building elements and materials. BIM and material passports can organize this information to support decisions about retention, reuse, and recycling \cite{finamore2026circularity}. However, creating reliable BIMs for existing buildings requires substantial manual effort, and suitable digital records are often unavailable. Reconstructing structured building models from photographs could reduce this effort by identifying architectural components, separating them from temporary objects, and representing partially occluded elements. Enriched with verified material and condition information, these models could support maintenance, renovation, adaptive reuse, and end-of-life recovery. This motivates our focus on hierarchical reconstruction as a step towards image-based BIM generation.
% Moreover, a hierarchical and semantically rich 3D model is very close to an actual Building Information Model (BIM), and generating a hierarchical model is thus a good first step towards BIM. BIM is important since the global construction industry accounts for approximately $37\%$ of global emissions and $50\%$ of material extraction \cite{UNEnvironmentProgramme}. Accurate 3D building models are essential in determining whether a building can be repurposed instead of being demolished, thereby saving significant amounts of $\text{CO}_2$ emissions. Even in cases where a building must be condemned, a BIM could help determine what materials, such as bricks, we might be able to salvage.

To produce such models, ARCH3D extracts geometric and semantic constraints from input photographs to guide the agent in constructing a geometrically accurate building model.
In summary, our contributions are: 
%\begin{itemize}
%    \item An AI agent which can make measurements based on the geometric information extracted from the photographs
%\end{itemize}

%--------- \textbf{We choose one or make a better one} ----

\begin{itemize}
    \item A fully automatic pipeline for reconstructing hierarchical editable 3D building models from photographs.
    \item A workflow that integrates semantics-aware geometry extraction with agent-guided measurement, modelling, and iterative refinement.
    %\item Structured model outputs with editable architectural components and functional assemblies, supporting batch operations and further refinement.
    \item An evaluation of how harness control, model choice, and reasoning effort affect reconstruction accuracy and computational cost.
\end{itemize}
\section{Related work}
\label{sec:Related_work}

The literature on 3D shape reconstruction from a set of photographs is vast, and a comprehensive overview is out of scope for this paper. In the following we thus outline the different research directions, which focus on creating a 3D shape from photographs directly, or from which the shape can be obtained as a by-product.

\textbf{Novel view synthesis methods:} In the very recent past, a common way to go from photographs to a 3D representation would be through novel view synthesis methods like the pioneering works NeRF \cite{mildenhall2020a} and 3D Gaussian Splatting \cite{kerbl2023a}, and their follow-ups \cite{muller2022a, fridovich-keil2022a, chen2022a, barron2023a, yu2024a_mip_splatting}. These methods are concerned with creating a scene representation either implicitly or explicitly from a set of input photographs. This allows for generating new views of the scene, which have not been captured by the photographs. While visually accurate, the focus of our method is not novel view synthesis but instead the underlying 3D geometric shape. %A surface in the form of a polygonal mesh can indeed be extracted from these scene representations using Marching Cubes \cite{lorensen1987a} for the implicit representation, or one of the following methods \cite{guedon2024a, huang2024a, yu2024a} for a Gaussian Splatting based representation, but a polygonal mesh is inherently not a hierarchical and semantically rich 3D model, which is the output of our methods.

\textbf{Mesh generative methods:} This research area covers many different works. Naturally, a dense point cloud can be obtained from a set of photographs using COLMAP \cite{pan2024glomap} or VGGT \cite{wang2025vggt}, and a surface can then be reconstructed using e.g. Poisson surface reconstruction \cite{kazhdan2006a} or other methods \cite{groueix2018a, rakotosaona2021a}. An example of a work, which directly generates a 3D shape from a set of photographs, is NeuS \cite{wang2021a}. A Signed Distance Function is learned for an object depicted in the photographs, and from the zero-level set, a polygonal mesh can be extracted. Another example of an image-conditioned 3D shape generation work is Trellis.2 \cite{xiang2025trellis2}. In this work, the authors introduce a new voxel structure, which they combine with a flow matching model to map from images to 3D meshes. Besides a mesh, their method outputs appearance and material information, which our model can also readily incorporate. All of these methods produce a mesh as an output, but a mesh is inherently not a hierarchical and semantically rich 3D model, which is the output of our methods. %Although these attributes enrich the mesh, they do not provide the individually manipulable components available in our hierarchical model.

%\textbf{Diffusion based methods:} Another line of research focuses on generating 3D shapes using diffusion models conditioned on images of the object and/or text-descriptions. In DreamFusion \cite{poole2023a}, the authors train a Diffusion model to generate images from a text prompt, which is then combined with a 3D scene parametrization model similar to NeRF. Based on this, they are able to extract a polygonal mesh. In TetraDiffusion \cite{kalischek2025a}, the authors propose a method based on a tetrahedral grid, for which they define a convolution operation. This allows them to apply a diffusion process to the tetrahedral grid, and thus generate 3D shapes as polygonal meshes.

%\textbf{Autoregressive methods:} Recently, there has also been an increased interest in generating polygonal meshes directly using transformer based methods, which generates meshes auto regressively as seen in MeshGPT, MeshArt \cite{gao2025a} , 

\textbf{CAD generating methods:} A CAD model has many of the same benefits as our hierarchical representation. Within the realm of CAD generative models, the authors of \cite{doris2025a} proposed CAD-Coder. It consists of a fine-tuned LLaVA 1.5 VLM, which can generate a parametric model of a mechanical part based on an image of it, by outputting a CADQuery Python script. Another work within this field is CAD-Assistant \cite{mallis2025cad}. In contrast to CAD-Coder, CAD-Assistant is a training-free VLM framework, which uses a set of pre-defined tools to assist a human CAD-modeller in answering queries about the CAD-design like the number of visible holes, and turn an image into a parametric model. Although similar in spirit, the above works only focus on simple mechanical components, whereas our method focuses on entire buildings.

\textbf{Agentic methods:} One of the first works within this domain is VIGA \cite{yin2026a}. In this paper, the authors demonstrate how a multimodal agent can generate a 3D model of an object based on sparse single-view inputs, which can be either photo-realistic images or hand-drawn sketches \cite{yin2026a}. Based on the input image and a prompt, the agent formulates a plan on how to achieve the goal, and then synthesizes python code in Blender to create the 3D model. With the rendered images of the model and a set of tools that the agent is equipped with, it can inspect the model and improve it iteratively. While similar in spirit, our method targets a different reconstruction representation from VIGA. VIGA uses external models such as Meshy \cite{meshy} to generate polygonal assets that match objects in the input image. Our method constructs a hierarchical parametric building model, in which architectural elements are represented as individually addressable components. Building on the same principles, Articraft \cite{zhou2026articraft} and LEGO-Anything \cite{li2026a} use AI-agents, which generate objects or scenes through iterative optimization based on input images or text prompts. Of the two, LEGO-Anything is closest to our approach, as it uses segmentation masks and depth estimations to generate a 3D scene. Another highly related work is AstraLOD3 \cite{pantoja-rosero2026a}. The authors use GPT-6 Astra \cite{openai2026astra} to generate Python scripts that construct 3D models in Blender \cite{Blender}, an open-source 3D creation application. However, their agent operates with lightweight geometric data and a limited set of reconstruction tools. Our method advances this approach by grounding model construction in explicit geometric measurements and verification across multiple views, enabling more precise control over building dimensions, proportions, and element placement.

%  VIGA is thus training-free, as the agent is powered by VLMs like Qwen3-VL-8B or GPT-4o. While similar in spirit, our method differs in terms of output. VIGA uses external models like Meshy \cite{meshy} to generate 3D assets, which are polygonal meshes, in order to match the object in the input image. Our method generates hierarchical parametric model instead.

%-------------------------------------------------------------------------
\section{Method}
\label{sec:Method}

\begin{figure*}[bt]
    \centering
    \includegraphics[width=\textwidth]{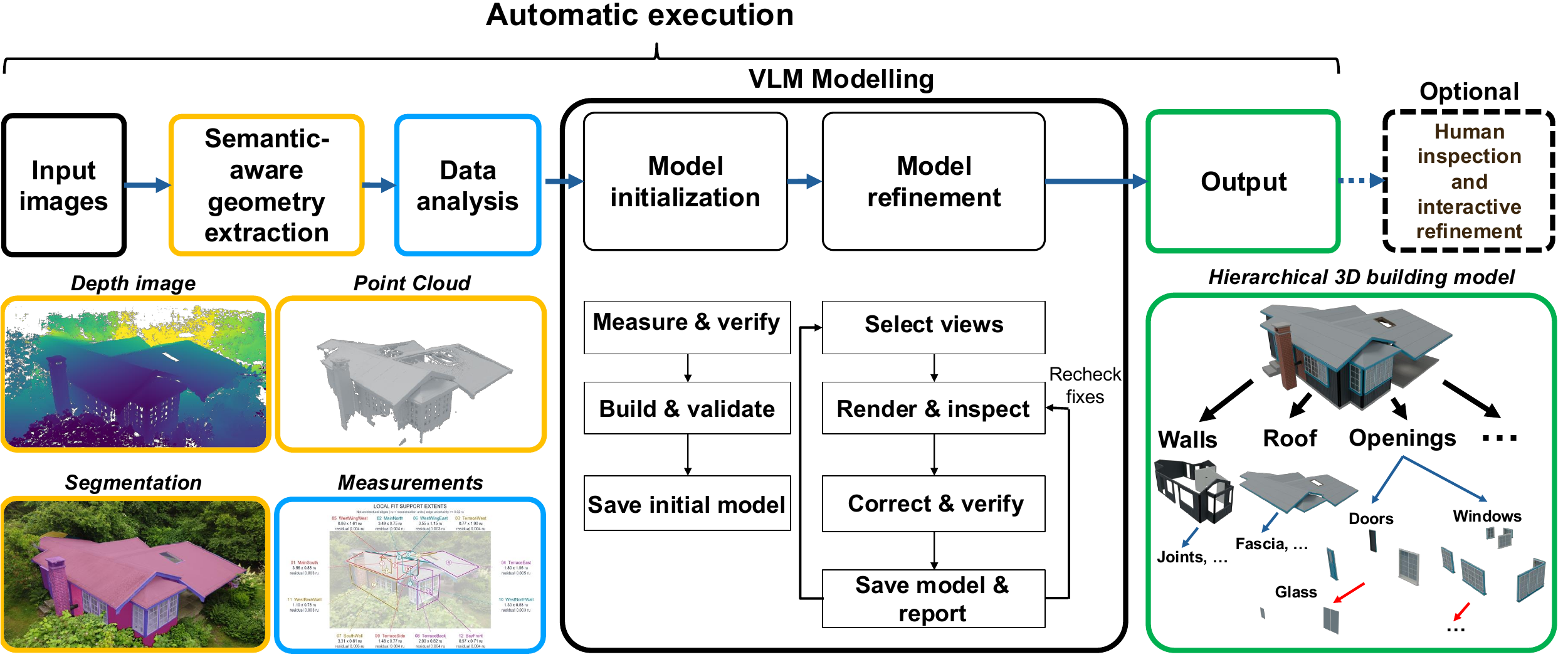}
     \caption{Pipeline of ARCH3D (best viewed digitally). Our method automatically reconstructs a geometrically accurate and hierarchical 3D building model from the input images. We achieve this by instructing the agent to obtain measurements from the geometric and semantic information extracted from the images and use the measurements, when modelling the building. Users can inspect and refine the model directly in Blender or request further revisions by continuing the dialogue with the agent.}
\label{fig:pipeline}
\end{figure*}

ARCH3D takes a set of photographs of a target building as input and generates a hierarchical 3D model, as illustrated in \cref{fig:pipeline}. The pipeline begins with semantic-aware geometry extraction, which provides explicit geometric and semantic information to the agent. The agent then analyzes this evidence to measure the dimensions, proportions, and positions of semantic building elements. During the subsequent modelling stage, it uses these measurements together with all available evidence to construct a hierarchical geometric 3D model of the building.

% With semantic geometry extraction, the agent can utilize geometric and semantic information extracted from the input images during modelling. The extracted information includes camera poses, a dense point cloud with semantic information, depth images, and instance-level segmented images. 
% We further investigate how reducing the Level of Control (LoC) affects reconstruction quality. Specifically, LoC-2 and LoC-1 progressively delegate more of the reconstruction process to the agent, as shown in \cref{fig:LoC_description}.

% ARCH3D takes a set of target building images as input and generates a hierarchical 3D model of the building. Depending on the using scenarios, users can choose different Level of Control (LoC) for different purposes. 

%To ground the output of the agent in physical constraints, we pre-process the input images to extract geometrical information such as camera poses, dense point clouds, depth images and segmented images. Depending on the Level of Control (LoC), explained below, the agent can utilize  %our method first generates a dense point cloud from the images using COLMAP \cite{pan2024glomap}. We then use SAM 3 \cite{carion2025sam3segmentconcepts} to segment the images and use these to filter out the points in the point cloud, which do not belong to the building. From the filtered 

%-------------------------------------------------------------------------

\subsection{Semantic-aware geometry extraction}

Semantic-aware geometry extraction begins with COLMAP \cite{pan2024glomap}, which estimates camera intrinsics and poses, reconstructs sparse and dense coloured point clouds, and produces undistorted images with corresponding depth maps. We then apply SAM3 \cite{carion2025sam3segmentconcepts} to the undistorted images using predefined prompts, retaining both prompt-specific and individual instance masks, along with confidence scores, bounding boxes, and depth statistics. These outputs preserve pixel-wise alignment between image appearance, depth, and semantic labels. To filter and semantically label the point cloud, we assign each prompt a unique ID and compute the union of all prompt masks in each image. We restrict stereo fusion to pixels within these unions, retaining points supported by the segmented regions. Each resulting 3D point is then projected into the reference images and assigned a semantic label based on the prompt IDs of the corresponding masks. We further align the geometry by rotating the estimated vertical direction onto the Z-axis, aligning the longest reliable fitted wall with the X-axis, and translating the bottom-center of the robust bounding box to the origin. Fitted plane equations and the bounding box descriptions are exported as structural references. The corresponding world-coordinate transformation is also applied to the camera poses. We then collect all available evidence in a structured data package. Each retained image is stored in a dedicated subdirectory together with its corresponding depth map, segmentation results, and camera metadata. Scene-level geometric information, including the bounding box and fitted planes, is stored at the package root.

\subsection{Agentic modelling}

Given the evidence package including semantic and geometric information, the agent is instructed to measure geometric features using the dense point cloud with semantic information, and visualizes the measurements through projection onto reference images. With this evidence, the agent starts modelling the target building.

The modelling stage is divided into two parts: \textit{initialization} and \textit{refinement}. During initialization, the agent inspects all provided information, and constructs a complete parametric Blender model. It then validates the model against the available evidence and corrects supported errors to deliver an initial model. Refinement proceeds for at least $N_r$ rounds, each using $N_i$ previously unselected, complementary reference views, or fewer when insufficient suitable views remain. We set $N_r=5$ and $N_i=5$ in all experiments. Within each round, the agent renders the model using the COLMAP-estimated camera intrinsics and poses of the selected views and compares the renderings with the corresponding reference images. It also inspects aligned point-cloud overlays with a translucent model, together with rendered and reference depth maps. Guided by this evidence, the agent corrects discrepancies in order of geometry soundness, major structure, and architectural details. It then rerenders the revised model from the same cameras, using depth error, semantic mask IoU, and point-cloud-to-model distance to support verification. Further corrections occur within the round before the agent saves the complete model and report. The prompts prescribe the operation order and refinement budget, while the agent determines view selection, evidence interpretation, and modelling operations. 

% \begin{figure}[]
%     \centering
%     \includegraphics[width=0.8\linewidth]{figures/workflow.png}
%      \caption{LoC-3 workflow for model initialization and iterative refinement.}
% \label{fig:workflow}
% \end{figure}

% \textbf{LoC-4: algorithmic pipeline specification}

\subsection{Agentic prompting: an analogy to numerical optimization}

We interpret agent-guided reconstruction by analogy to iterative numerical optimization. The building model represents the current solution, and the agent acts as the optimizer. In each refinement round, the agent compares renderings of the model with reference data and uses the observed discrepancies to propose an update. The prompt specifies the optimization objective and hyperparameters. It guides the choice of update and its scope, corresponding to the update direction and step size in gradient descent, respectively. Users can tune these hyperparameters through the prompt qualitatively. The initialization also matters. A plausible initial solution can reduce the number of rounds needed for structural correction.

Our experiments reveal six limitations of the agent as an optimizer, which inform our prompt design, workflow control, and experimental protocol. (a) With long prompts, the agent can overlook requirements or fail to apply them in later rounds. We therefore maintain persistent context and modelling requirements in \texttt{Context.txt} and \texttt{Contract.txt}, respectively, and instruct the agent to reread both files before each stage. (b) The agent can generate the model before producing the analysis outputs intended to guide its construction. Consequently, explicit operation sequences address this issue by requiring analysis before model generation or revision. (c) The agent can make overly conservative updates, addressing only one issue despite identifying multiple discrepancies. Therefore, to encourage comprehensive corrections, we instruct it to address all actionable, evidence-supported discrepancies in each update. (d) When allowed to determine termination, the agent can stop before sufficiently diagnosing and correcting remaining errors. To combat this, our prompts prescribe a refinement budget, stopping criteria, and evaluation metrics to guide continued diagnosis and correction. (e) Natural-language prompts can leave workflow details open to interpretation. This motivates our investigation of algorithmic control over the workflow of LoC-3, which encodes the workflow algorithmically to reduce procedural ambiguity. (f) Modelling quality can vary across runs, so we average results over repeated trials to reduce the influence of this variability. The prompts are available in the supplementary material.

%-------------------------------------------------------------------------

%\subsection{Real houses}

%\subsection{TwinTex}

%\subsection{Minneapolis commercial buildings}

%\subsection{NYC dataset}

%-------------------------------------------------------------------------
\section{Experimental settings}

We use Codex \cite{openai2026Codex} as the agent harness for ARCH3D, with GPT-6 Astra \cite{openai2026astra} serving as the underlying model in most experiments. Our framework specifies the reconstruction tasks and prompts, which the agent executes through the harness.

\subsection{Datasets}\label{subsec:datasets}

We evaluate our method on four datasets: Ten buildings from \textit{NYC}, a modified synthetic building dataset derived from the Valdrada project \cite{turkcan2026valdrada}, which is a digital twin of New York City’s five boroughs; two buildings from \textit{Matt1up} \cite{guertin2026drone}, scan data and models of two commercial buildings in Minneapolis; five buildings from \textit{TwinTex} \cite{TwinTex23}, drone imagery of real high-rise buildings, from which we select five of the ten scenes depicting individual building exteriors, excluding scenes containing interiors or groups of buildings; and 3 buildings from \textit{Cottages}, our own drone footage of cottages with distinct architectural styles.

For \textit{NYC}, we procedurally assign random surface colours to each model to reduce repetitive appearance patterns and facilitate COLMAP \cite{pan2024glomap} reconstruction. This preprocessing compensates for the limited appearance variation in the synthetic models, whereas real-world photographs typically provide richer visual cues.

\subsection{Execution environment}

We use Harbor \cite{harbor}, a framework for running and evaluating AI agents in isolated task environments, to execute each experiment as a separate Docker-based trial. Each trial starts with a fresh workspace and Codex session to prevent state from carrying over between experiments. Input resources were mounted read-only at \texttt{/app/Resources}, while prompts and workflow code were snapshotted for each trial. Each Docker container provides Blender 4.5.15 LTS, Python dependencies, and CuPy, with NVIDIA GPU access for rendering and depth computation. Trials ran sequentially to avoid GPU contention. A specific trial can be resumed by specifying its saved job directory.

\subsection{Evaluation metrics}

We evaluate the 3D building models generated by ARCH3D and the baseline method, AstraLOD3, using the 2D Intersection over Union (\textbf{2D IoU}) and the point cloud to surface distance \textbf{pc-Dist}.% and 3D IoU (\textbf{3D IoU}). %The last metric is however only application for the \textit{NYC} dataset, where ground-truth 3D building model is provided. The results are summarized in \cref{tab:Results}. 

\begin{figure}[]
    \centering
    \includegraphics[width=\linewidth]{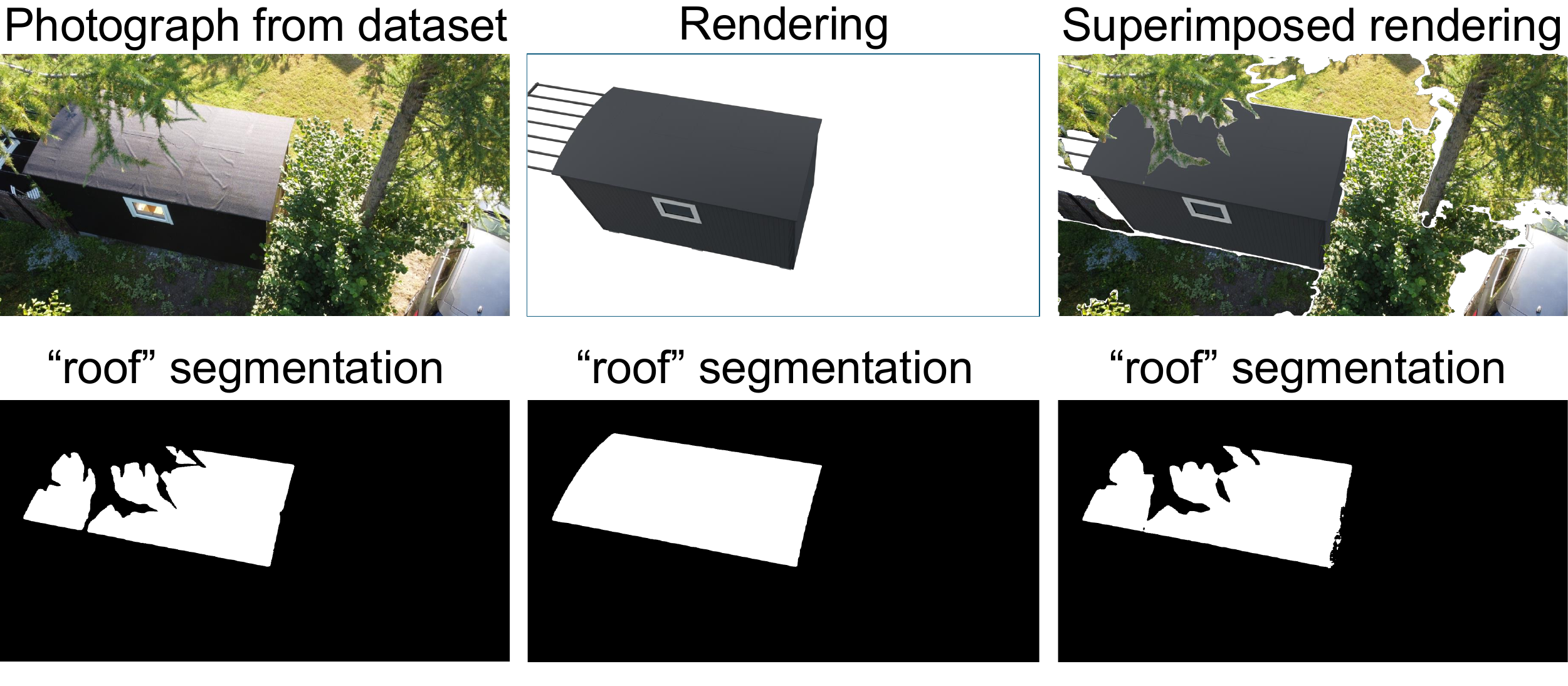}
     \caption{Segmenting the dataset photograph (\textit{cottage dataset}) using SAM3 and the prompt "roof" will give a small segmentation mask (left column) due to the occlusion from the tree. The segmentation mask computed in the same way for the rendered image (middle) will however be bigger. We therefore superimpose the surroundings on top of the rendering and segment the superimposed image to get a segmentation mask (right column), which matches the photograph segmentation. The images are cropped in the bottom to leave out the drone-pilot}
\label{fig:segmentation_masks_2D_IoU}
\end{figure}

To compute the \textbf{2D IoU} measure, we render images of the agent-generated 3D building model in Blender using the camera intrinsics and poses obtained from running COLMAP \cite{pan2024glomap} on the dataset photographs. This allows us to do a pairwise comparison between a photograph of a building in the dataset and its corresponding rendered image. For each \{photograph, image-rendering\}-pair, we use SAM3 to compute segmentation masks of the structural elements (see supplementary material for the prompts), and based on these masks we can compute the \textbf{2D IoU}. This approach is however challenged by the fact that the photographs may contain separate buildings, and non-building-related objects like vegetation, outdoor-furnitures, cars etc., which can occlude the view of the building in question. This occlusion can cause the segmentation mask of the photograph to be smaller than the corresponding mask in the rendered image, as the agent-generated building did not include objects, which could occlude it in the renderings. This difference is illustrated in \cref{fig:segmentation_masks_2D_IoU}, and will naturally lead to a lower IoU score, despite a correct segmentation mask from the rendering of the geometrically accurate building. Furthermore, SAM3 could be affected by the surrounding objects, when generating the segmentation mask from the photographs. To counteract these problems, we thus use SAM3 to first segment objects unrelated to the building in question in the photographs, and superimpose these objects on the rendered images of the agent-generated building. Afterwards, we use SAM3 to segment the structural elements in the superimposed renderings, so we now measure the \textbf{2D IoU} between the segmentation masks of a photograph and its corresponding superimposed rendered image. This approach can however allow for SAM3 to incorrectly segment parts of the building in the photographs, which will then be superimposed on top of the renderings. For each dataset \textbf{2D IoU}, we report the mean $\mu$, and the 25th, 50th and 75th percentiles ($P_{25}, P_{50}, P_{75}$) respectively.
We measure the point cloud-to-surface distance (\textbf{pc-Dist.}) by creating a bounding volume hierarchy of the generated 3D model using Blender, after which we compute the smallest distance from each point in a filtered dense point cloud generated by COLMAP to the 3D model. The filtering process of the dense point cloud from COLMAP consisted of two stages. First, the segmentation masks from SAM3 were used to reject points not related to the building. Afterward, manual filtering was applied to remove the remaining non-building-related points which passed the first stage. Similar to the \textbf{2D IoU} measure, we report the mean, $\mu$, and the 25th, 50th, and 75th percentiles ($P_{25}, P_{50}, P_{75}$) respectively, of the distances measured in Reconstruction Units (RU), as we do not have access to ground truth distances. 
%For the computation of \textbf{3D IoU}, we register the synthetic buildings to the dense point clouds obtained from COLMAP. For each synthetic building we generate an evenly spaced voxelgrid of resolution $128^3$ containing the synthetic buildings, and we then compute the Generalized Winding Number \cite{jacobson2013a} at each voxel point using libigl \cite{libigl} to asses, whether a voxel is inside a generated building and a synthetic building. Based on the ratio between voxels inside and voxels outside, we compute the 3D IoU.

\begin{figure*}[]
    \centering
    \includegraphics[width=0.9\textwidth]{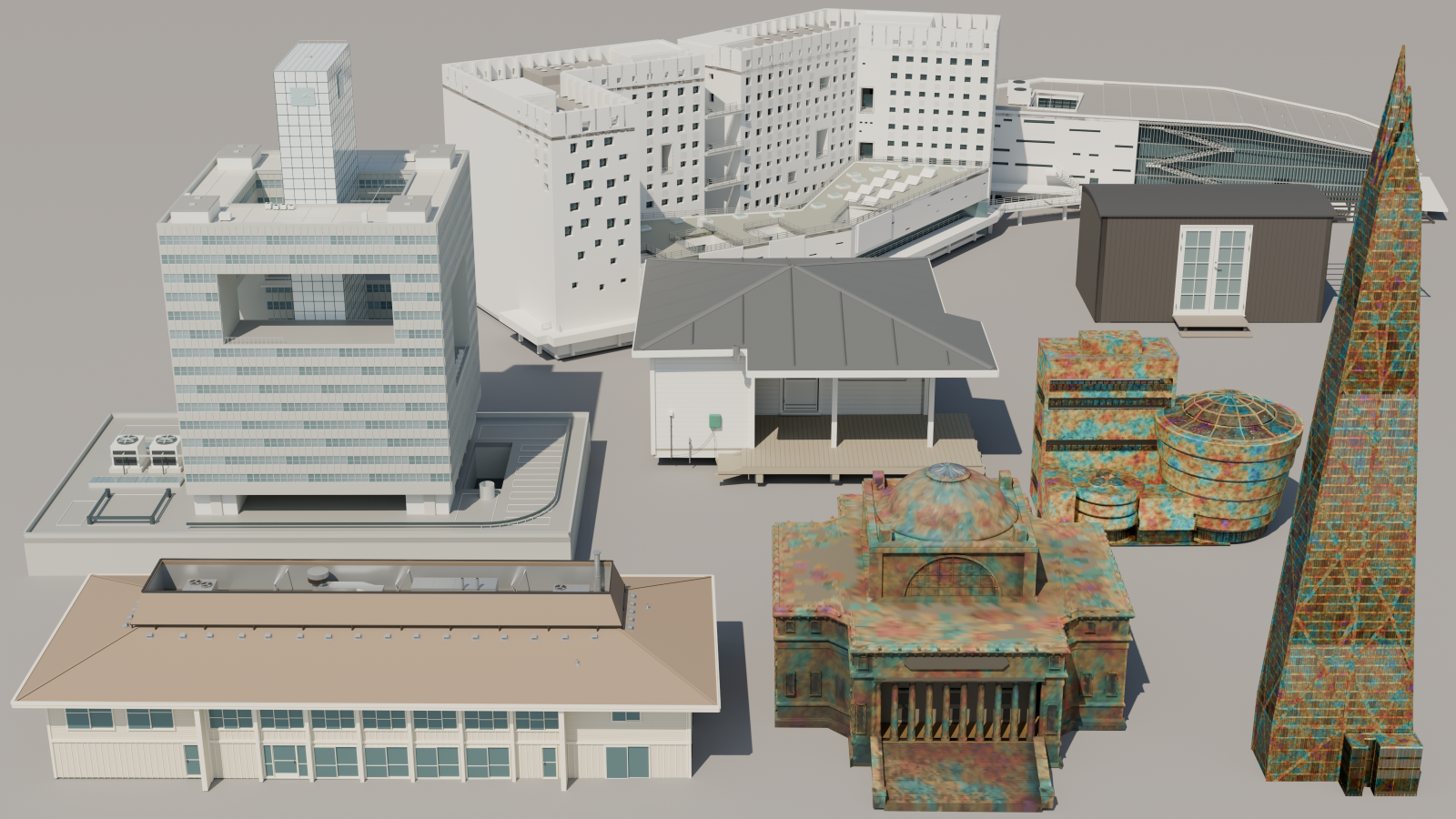}
     \caption{A gallery of selected models from all four datasets reconstructed by ARCH3D. For synthetic buildings with repetitive structures (\textit{NYC} dataset) which can challenge the SfM from COLMAP, we overlay the buildings with a new texture.}
\label{fig:gallery}
\end{figure*}
\begin{figure*}[]
    \centering
    \includegraphics[width=\textwidth]{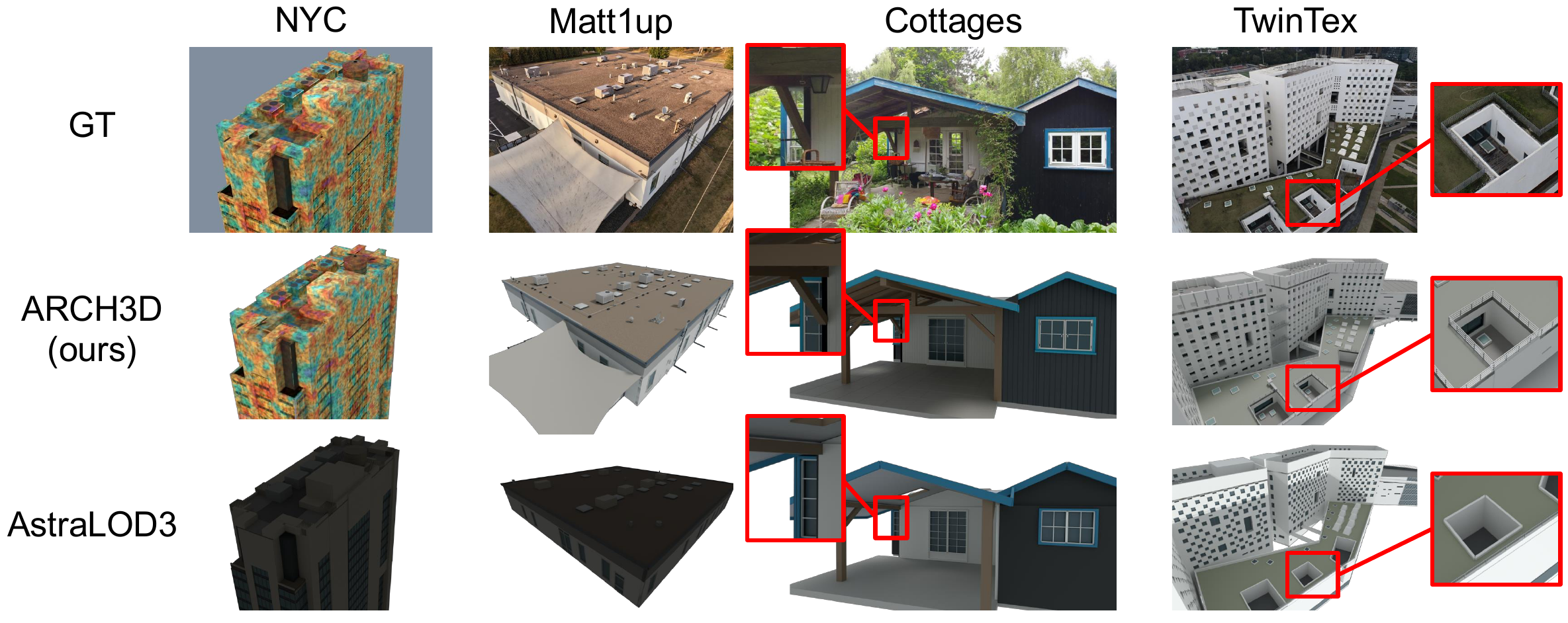}
     \caption{Qualitative comparison of the 3D building models generated by ARCH3D and AstraLOD3}
\label{fig:comparison}
\end{figure*}

% Some models appear darker because their meshes were exported from separate Blender files and rendered under shared lighting, which does not reproduce each model’s original lighting conditions.

\section{Results}

We first present a qualitative overview of the reconstructed models in \cref{fig:gallery}. All results of ARCH3D are directly from the automatic execution output, without any human inspection and interactive refinement. Complete visualizations of all reconstructed models are provided in the supplementary material.

\subsection{Comparisons across datasets}

We evaluate our method quantitatively and qualitatively across all datasets. To keep runtime and token costs manageable, we conduct one trial per scene for the comparative evaluation. \cref{fig:comparison} shows representative reconstructions from each dataset, with enlarged views highlighting differences in architectural details. Quantitative results in \cref{tab:Results} show that ARCH3D outperforms the baseline in most evaluations. For the \textit{TwinTex} dataset we observe that our ARCH3D outperforms AstraLOD3 in terms of geometric reconstruction, but in terms of \textbf{2D IoU} both methods perform poorly. This can be attributed to the fact that the photographs in this dataset contain multiple buildings, and it is therefore difficult to use SAM3 to only segment the building in question and its structural elements.

It should be noted that SAM3-based comparisons are influenced by the rendering quality, and SAM3 can produce inaccurate or incomplete segmentations on both photographs and rendered images. However, relative differences under a consistent evaluation protocol are informative.
\subsection{Ablation studies}

To evaluate our design choices, we conduct an ablation study, where we investigate the performance of our pipeline with respect to the different Levels of Control (LoC), the models behind the AI-agent, and the reasoning level on the cottage in \cref{fig:teaser} from the \textit{Cottages} dataset. For each experiment setting, we run $5$ trials and report the \textbf{2D IoU} as well as the point cloud distance \textbf{pc-Dist.}. We summarize the results in \cref{tab:Ablation_study}.  

%\subsubsection{Level of control (LoC) for agent-guided parametric modelling}\label{subsec:LoC}

\begin{figure}[]
    \centering
    \includegraphics[width=\linewidth]{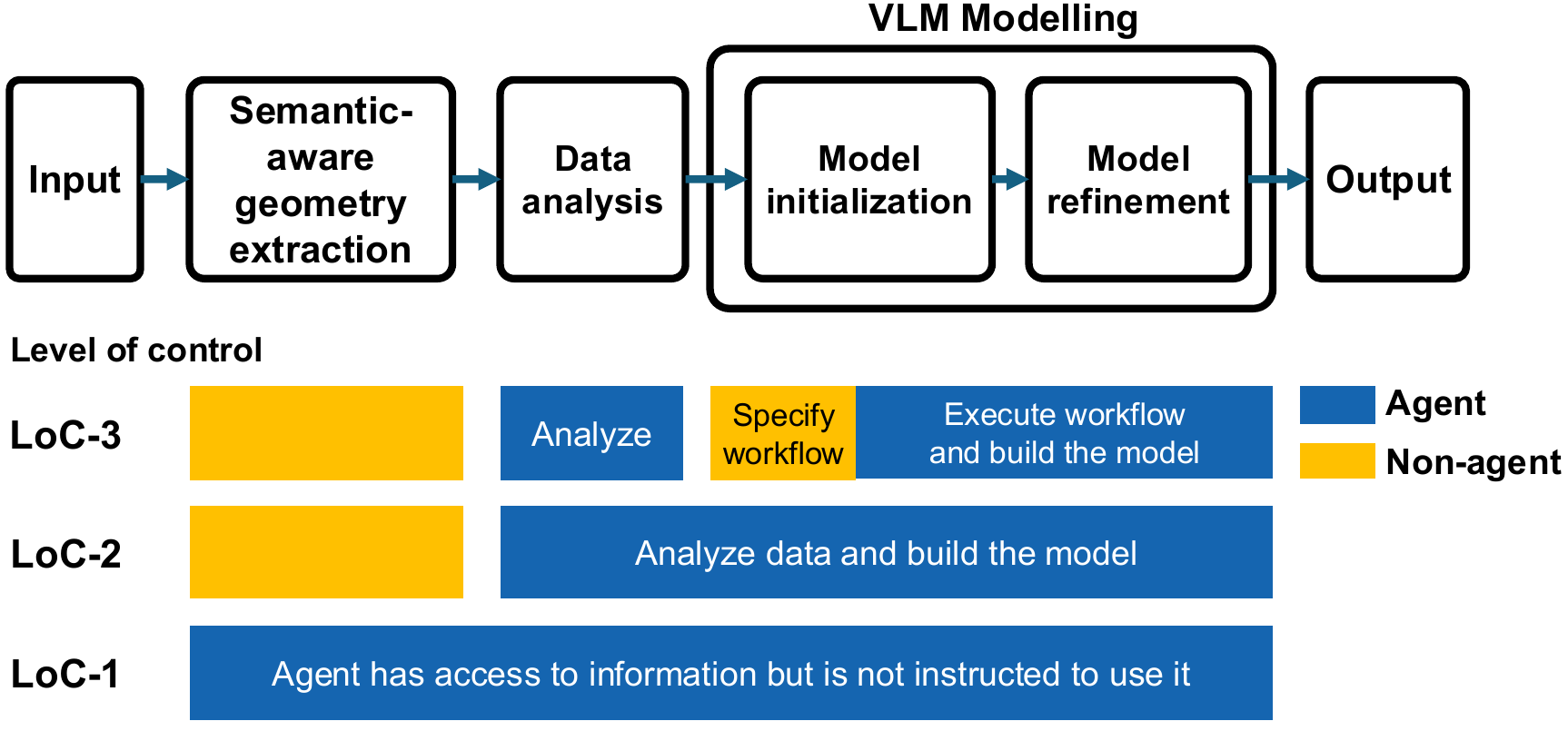}
     \caption{3D building model reconstruction pipeline and levels of control (LoC).}
\label{fig:LoC_description}
\end{figure}
% LoC-1 assigns all intermediate stages to the agent. LoC-2 introduces human data preprocessing, and LoC-3 additionally assigns workflow definition to the human.

We investigate three LoCs, distinguished by the extent to which the harness constrains the agent’s actions and workflow. A graphical representation of the different LoCs is provided in \cref{fig:LoC_description}. 

\textbf{LoC-1: full autonomy} In the lowest LoC-1, the agent has full autonomy and is tasked with generating the 3D building model solely from the input photographs. 

\textbf{LoC-2: input resource preparation} At LoC-2, we further process the source photographs using structure from motion (SfM) and semantic segmentation to extract geometric and semantic information. The agent receives these outputs alongside the source photographs and independently determines the modelling workflow.

\textbf{LoC-3: agent workflow specification} Based on LoC-2, we further tighten the harness of the agent by providing a pre-defined workflow.

As shown in \cref{tab:Ablation_study}, tighter harness control improves reconstruction accuracy up to LoC-3. For the LoC-1 trials, we registered the 3D models to the dense point clouds using a manual alignment followed by ICP, as the agent could not place the 3D building model close to it without access to a dense point cloud, leading to erroneous results. 

% We further investigate LoC-3 with algorithmic constraints, which impose algorithmic control over the workflow, and find that excessive constraints have diminishing returns. A detailed comparison is provided in the supplementary material.

\begin{figure}[t]
    \centering
    \includegraphics[width=\linewidth]{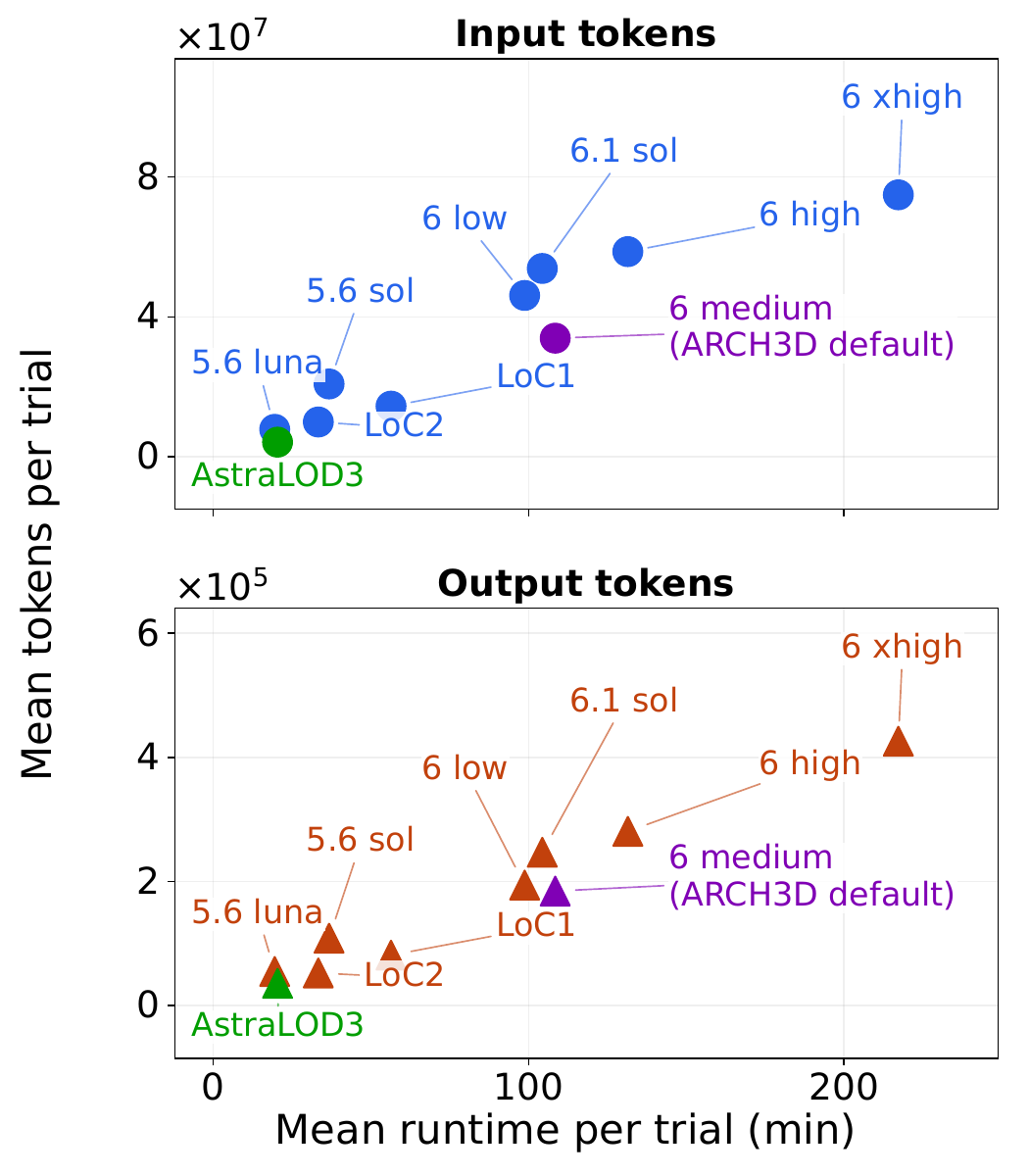}
     \caption{Average execution time and token consumption per trial on the cottage in \cref{fig:teaser} (\textit{Cottage} dataset) for AstraLOD3 and ARCH3D across levels of control (LoCs), reasoning effort levels, and model variants. Each point represents the mean over five trials. The top and bottom panels show input and output token consumption, respectively; the horizontal axis shows mean execution time per trial.}
\label{fig:time_and_token}
\vspace{-3mm}
\end{figure}

\textbf{Reasoning effort and model variants} We further evaluate how reasoning effort and GPT model variants affect reconstruction quality. The last two sections of \cref{tab:Ablation_study} present the results. Pre-Astra models generally failed to closely reproduce the photographed building. Increasing reasoning effort does not always substantially improve reconstruction quality, as illustrated by the low and medium settings. Comparing these results with the execution time and token consumption in \cref{fig:time_and_token} helps assess whether improvements in reconstruction quality justify the additional computational cost.

\begin{table*}[ht]
  \centering
  \setlength{\tabcolsep}{10pt}
  \caption{\textbf{Results:} We compare our method against AstraLOD3 on four different datasets \textit{NYC} \cite{turkcan2026valdrada}, \textit{Matt1up} \cite{guertin2026drone}, \textit{TwinTex} \cite{TwinTex23} and \textit{Cottages}. $\uparrow$ indicates higher is better, and $\downarrow$ indicates lower is better. The best performance is highlighted in bold.} %N/A means Not Applicable, as \textbf{3D IoU} can only be computed for \textit{NYC} with ground truth 3D models}
  \begin{tabular}{@{}cccc@{} }
    \toprule
    \textbf{Dataset} & \textbf{Method} & \textbf{2D IoU} [$\mu,P_{25},P_{50},P_{75}$] $\uparrow$    & \textbf{pc-Dist.} [$\mu,P_{25},P_{50},P_{75}$] $\cdot 10^3$ $\downarrow$   \\
    \midrule
    \multirow{2}{*}{\textit{NYC} \cite{turkcan2026valdrada}}
        & AstraLOD3 \cite{pantoja-rosero2026a}
        & $0.647/0.407/0.720/0.964$
        & $5.48/0.821/2.35/6.20$ \\
        & ARCH3D (ours)
        & $\mathbf{0.726}/\mathbf{0.585}/\mathbf{0.837}/\mathbf{0.976}$
        & $\mathbf{1.88}/\mathbf{0.304}/\mathbf{0.811}/\mathbf{1.99}$\\
    \midrule
    \multirow{2}{*}{\textit{Matt1up} \cite{guertin2026drone}}
        & AstraLOD3 \cite{pantoja-rosero2026a}
        & $0.377/0.0/0.206/0.813$
        & $27.0/1.87/4.21/8.74$\\
        & ARCH3D (ours)
        & $\mathbf{0.481}/0.0/\mathbf{0.633}/\mathbf{0.861}$
        & $\mathbf{5.96}/\mathbf{0.851}/\mathbf{1.98}/\mathbf{3.98}$\\
    \midrule
    \multirow{2}{*}{\textit{TwinTex} \cite{TwinTex23}}
        & AstraLOD3 \cite{pantoja-rosero2026a}
        & $0.203/0.0/0.0/\mathbf{0.447}$
        & $17.0/2.55/5.86/13.0$ \\
        & ARCH3D (ours)
        & $\mathbf{0.206}/0.0/0.0/0.435$
        & $\mathbf{5.74}/\mathbf{1.06}/\mathbf{2.82}/\mathbf{6.33}$ \\
    \midrule
    \multirow{2}{*}{\textit{Cottages}}
        & AstraLOD3 \cite{pantoja-rosero2026a}
        & $0.436/0.0/0.510/0.819$
        & $18.8/4.94/12.3/24.8$ \\
        & ARCH3D (ours)
        & $\mathbf{0.594}/\mathbf{0.233}/\mathbf{0.763}/\mathbf{0.905}$
        & $\mathbf{8.35}/\mathbf{1.49}/\mathbf{3.93}/\mathbf{9.68}$ \\
    \bottomrule
  \end{tabular}
  \label{tab:Results}
\end{table*}

%Evaluation metrics:

%2D Mask IoU, point cloud to surface distance, 3D IoU with synthetic model / Chamfer distance between reconstructed model with synthetic model.

%-------------------------------------------------------------------------

\begin{table*}[t]
  \centering
  \setlength{\tabcolsep}{10pt}
  \caption{\textbf{Ablation study:} We evaluate the performance of our framework with respect to different design choices such as Level of Control (LoC), the VLM model, and the reasoning level of the VLM. We evaluate these design choices on the cottage in \cref{fig:teaser}. from the \textit{Cottages} dataset. $\uparrow$ indicates higher is better, and $\downarrow$ indicates lower is better. The best performance within each category is highlighted in bold. To account for the inherited randomness, we run 5 trials for each model. N/A means Not Applicable, as LoC-1 does not require the 3D building model generated by the VLM to be coherent with the input images}
  \begin{tabular}{@{}lcc@{} }
    \toprule
    \textbf{Method} & \textbf{2D IoU} [$\mu,P_{25},P_{50},P_{75}$] $\uparrow$  & \textbf{pc-Dist} [$\mu,P_{25},P_{50},P_{75}$] $\cdot 10^3$ $\downarrow$  \\
    \midrule  
    \multicolumn{3}{@{}l@{}}{\textit{Level of Control} (GPT-6 Astra / Reasoning effort: Medium):}\\
    \midrule
    LoC-1  & N/A & $66.4/10.3/29.8/68.1$   \\
    LoC-2  & $0.569/0.164/0.723/0.883$ & $11.9/1.96/5.09/12.1$  \\
    LoC-3  & $\mathbf{0.606}/\mathbf{0.304}/\mathbf{0.766}/\mathbf{0.900}$ & $\mathbf{8.80}/\mathbf{1.61}/\mathbf{4.30}/\mathbf{9.82}$  \\
    \midrule
    \multicolumn{3}{@{}l@{}}{\textit{GPT model} (Reasoning effort: Medium / LoC-3):}\\
    \midrule
    GPT-5.6 Luna & $0.112/0.0/0.0/0.0$& $150.7/26.8/82.8/239.6$ \\
    GPT-5.6 Sol & $0.257/0.0/0.0/0.550$ & $68.7/9.80/22.3/66.7$  \\
    GPT-6.1 Sol & $\mathbf{0.615}/\mathbf{0.347}/\mathbf{0.773}/\mathbf{0.89}$ & $\mathbf{7.90}/\mathbf{1.29}/\mathbf{3.19}/\mathbf{8.07}$  \\
    GPT-6 Astra & $0.606/0.304/0.766/0.900$ & $8.80/1.61/4.30/9.82$  \\
    \midrule
    \multicolumn{1}{@{}l@{}}{\textit{Reasoning effort} (GPT-6 Astra / LoC-3): }  \\
    \midrule
    Low & $0.607/0.308/0.772/0.903$ & $8.36/1.35/3.42/8.11$  \\
    Medium & $0.606/0.304/0.766/0.900$ & $8.80/1.61/4.30/9.82$  \\
    High & $0.635/0.421/0.787/0.905$ & $7.37/1.35/3.51/7.69$ \\
    Extra High & $\mathbf{0.658}/\mathbf{0.477}/\mathbf{0.815}/\mathbf{0.920}$ & $\mathbf{5.19}/\mathbf{0.809}/\mathbf{1.99}/\mathbf{4.88}$  \\
    \bottomrule
  \end{tabular}
  \label{tab:Ablation_study}
\end{table*}

\subsection{Time and token consumption}

We report execution time and token consumption across model variants and reasoning effort levels in \cref{fig:time_and_token}. All experiments use the cottage building depicted in \cref{fig:teaser} from the \textit{Cottages} dataset, with five independent trials per experimental condition. Together with the reconstruction accuracy reported in the preceding experiments, these results characterize the trade-off between reconstruction quality and resource usage, helping users select a configuration that suits their accuracy requirements, time constraints, and computational budget. As we would expect, increasing the reasoning level also increases the token-usage and runtime.

%\subsection{LoC}

%\subsection{Evidence provided to agent}

%\subsection{Agent models and reasoning level}

%-------------------------------------------------------------------------
\section{Discussion}
We have presented ARCH3D, a method that combines semantic-aware geometry extraction with agent-guided modelling to reconstruct hierarchical 3D building models from photographs. Its outputs contain editable architectural components and functional assemblies, supporting batch operations and refinement in Blender or through dialogue. We have studied the effects of level of control, model choice, and reasoning effort. We have also compared to the closest related effort, AstraLOD3 \cite{pantoja-rosero2026a}, and find that our reconstructions are uniformly more precise, albeit at greater cost in terms of both time and tokens. Unlike AstraLOD3, our method always produces lighting environments.
%We also observe improved lighting in the generated Blender scenes, which can influence rendered appearance and SAM3 segmentation quality, and thus the semantic evaluation.

%\subsection{Evaluation with SAM3: pros and cons}

%\subsection{Future work}

%-------------------------------------------------------------------------
%\section{Limitations}

The main limitations of our method are related to uniform repetitive structure, as exhibited by the synthetically created buildings in the \textit{NYC} dataset \cite{turkcan2026valdrada}. Moreover, details below a certain size (handles, knobs, plugs) tend to be captured but with less accuracy than architectural structure.

To conclude, ARCH3D provides a foundation for image-based BIM generation through hierarchical building reconstruction. Future work could integrate material properties and condition assessments into these models to support maintenance, adaptive reuse, and component recovery, advancing the broader goal of repurposing rather than demolition, and reducing construction waste.

% oreover, very small details cannot be captured. 
% Our method is challenged by buildings with uniform repetitive texture, which is the case for certain synthetically created buildings in the Valdrada dataset \cite{turkcan2026valdrada}. Our model is also challenged by modelling small details on buildings.

%-------------------------------------------------------------------------
%\newpage

\section{LLM Usage Statement}

This work proposes a framework to generate a 3D hierarchical building model from a set of photographs. The core component is an AI-agent, which relies on a Large Vision Language Model. During this project, we have also used AI-tools to assist with coding and for improving minor parts of the text.

\iffalse
\section{Appendix}

In the following we elaborate on our design choices.

\subsection{Computing the 2D IoU}\label{appendix:2D_masked_IoU}

When computing the 2D masked IoU, we use SAM3 to segment the images into 11 different pre-defined categories which are: "window", "door","roof","wall", "chimney", "garage","balcony","terrace","column","beam",
        "stair","building", "house", "porch", "canopy", "railing".

\subsection{Computation of distance metrics}

The symmetric Chamfer Distance is computed between a point cloud $S_1$ sampled uniformly on the 3D building model generated by the agent, and the point cloud $S_2$ sampled uniformly on the synthetic building model as: 
\begin{equation}
    d_{CD}(S_1, S_2) = \frac{1}{N} \Big ( \sum_{x\in S_1} \min_{y \in S_2} \lVert x -y \rVert_2^2 + \sum_{y\in S_2} \min_{x \in S_1} \lVert x -y \rVert_2^2 \Big )
\end{equation}
For the same point clouds $S_1$ and $S_2$, we compute the symmetric Hausdorff distance as: 
\begin{equation}
    d_{HD}(S_1,S_2) = \max \big (h(S_1,S_2), h(S_2,S_1)\big )
\end{equation}
where $h(S_1,S_2) = \sup_{x\in S_1} \inf_{y\in S_2}\lVert x-y\rVert_2$.
\fi

%-------------------------------------------------------------------------

% 
\clearpage

\setcounter{page}{1}
\maketitlesupplementaryonecolumn

The following text accompanies the paper "ARCH3D: Agentic Reconstruction of hierarCHical 3D building models" submitted for the Computer Vision and Pattern Recognition Conference 2027. The supplementary material will provide details on the implementation such as the prompts for ARCH3D and SAM3 in \cref{subsec:supp_ARCH3D_prompts} and \cref{subsec:SAM3_prompts} respectively, and present additional qualitative results on the datasets in \cref{sec:supp_3D_building_visualizaitons}. We will also report results from two sets of experiments, the first using a new level of control (LoC-3+) in \cref{subsec:supp_LoC3_add_introduction} and the second using vision-language models (VLMs) released after the paper submission deadline in \cref{subsec:New_GPT_models}. Furthermore, we provide pseudocode for semantic-aware geometry extraction in \cref{sec:supp_pseudocode_for_semantic_aware_geometry} and pseudocode for the new Level of Control (LoC-3+) in \cref{sec:supp_pseudocode_for_LoC3_add}. Finally, we discuss data availability in \cref{sec:supp_data_availability}. 

% \onecolumn
\section{Prompts}
\label{supplement:Prompts}

In the following we provide the prompts used in this work, including those for 3D building model reconstruction with ARCH3D, and semantic segmentation of photographs and images with SAM3 for evaluation. For the AstraLOD3 baseline, we use the original prompts from \cite{pantoja-rosero2026a} and refer readers to that work for details. The start of the prompt in \textit{Context} for ARCH3D is partly inspired by the prompts used in VIGA: Vision-as-Inverse-Graphis \cite{yin2026a}.

% In the following we will outline the different prompts which we used for our method, ARCH3D. This includes the prompts used with GPT-6 Astra to generate the 3D buildings, as well as prompts used with SAM3 to generate the segmentation masks.\\ 

% In our work, we compare ARCH3D to the baseline AstraLOD3. For AstraLOD3, we used the same prompts, which they used in their paper \cite{pantoja-rosero2026a}, and we therefore refer the reader to this work.

\subsection{ARCH3D}\label{subsec:supp_ARCH3D_prompts}

We use colors to distinguish prompts applied at each LoC: black for LoC-1, blue for LoC-2, and red for LoC-3. Each higher LoC incorporates the prompts from all lower levels. For example, the LoC-2 prompt includes both the black and blue text. Square brackets indicate alternatives, from which the entry corresponding to the selected LoC should be used. For instance, in \textbf{[R05 \textcolor{red}{R06}]}, we use R05 for LoC-2 and LoC-1, and R06 for LoC-3. Contract requirements are numbered consecutively from R01 within each LoC. Requirements omitted at lower LoCs do not leave gaps in the numbering. For example, if R04 is absent, R05 is renumbered as R04, and subsequent requirements are renumbered accordingly.

For LoC-3, the {context}, {contract}, {initialization}, and {refinement} prompts are stored in separate files and supplied sequentially by the algorithm, enabling automated execution while keeping each prompt concise.
\begin{Verbatim}[
    fontsize=\small,
    breaklines=true,
    breakanywhere=true,
    commandchars=\\\{\}
]
# Context

You are Builder3D, an expert in reconstructing parametric 3D building models through Python scripts for Blender.

## Objective
The goal is an identical digital twin of the target building: a precise reconstruction with correct overall geometry, scale, proportions, dimensions, orientation, and relative positions of all structural and architectural elements. This includes the building volumes, roof structure, facade planes, openings, doors, windows, frames, and architectural details, as well as their connections and intended door/window operation. The model should agree with the supplied evidence across all available views, not merely resemble the building from one viewpoint.

## Resource reference
Resource root: /app/Resources
\textcolor{blue}{Image package: /app/Resources/vlm_package}

\textcolor{blue}{Each image directory contains:}
\textcolor{blue}{- image.jpg: reference photograph.}
\textcolor{blue}{- depth.npy and depth_preview.png: depth array and visualization.}
\textcolor{blue}{- mask_{prompt_id}.png: prompt-specific semantic masks, with IDs 0=wall, 1=window, 2=door, 3=roof, 4=building, 5=house, 6=column, 7=beam, 8=balcony, 9=porch, 10=canopy, 11=stairs, 12=railing, 13=chimney, 14=garage. segmentation_overlay.png: colored overlay of the semantic masks on the undistorted RGB image.}
\textcolor{blue}{- instance_p{prompt_id}_i{instance_id}.png: individual instance masks.}
\textcolor{blue}{- metadata.json: camera settings, depth and segmentation information, and transformation matrices.}

\textcolor{blue}{The image package's manifest.json lists image names and folder organization. The resource folder also contains bbox.txt (rough axis-aligned building bounds), planes.txt (fitted point-cloud planes), and colmap_point_cloud.ply (dense cloud). All supplied resources share an aligned coordinate system.}

\textcolor{red}{This message supplies context only. Acknowledge and wait for Contract; do not begin work.}


# Contract

This contract governs setup, initialization, and every refinement round.

\textcolor{red}{[R01] Activation and requirement tracking}
\textcolor{red}{During setup, acknowledge Context and Contract and wait. Do not create code, modify files, generate analysis artifacts, or run Blender until the separate Initialization request. Refinement starts only upon a separate Refine request. On initialization, create the workspace and save this contract there as reconstruction_requirements.txt. Read it before modelling in initialization and each refinement round, and check R01-R10 before completing each round.}

[R02] Workspace and tools
Create \textbf{[}LoC1 \textcolor{blue}{LoC2} \textcolor{red}{LoC3}\textbf{]} under the caller's working directory. Keep all created or modified task files there, including scripts, models, renders, analysis, logs, and dependency environments/caches under your control. Resources are read-only. Do not inspect unrelated folders or projects; report missing data instead. Installed Python/Blender runtimes and required libraries may be accessed for execution. Use Python, Blender, and bpy; install additional libraries with pip when needed within these boundaries. Blender may be used for analysis, visualization, diagnostic renders, and testing after activation. \textcolor{blue}{Use the manifest's actual filenames, including zero-padded IDs.}

[R03] Evidence and reconstruction scope
\textcolor{blue}{Combine photographs, depth, semantic/instance masks, cameras, bounding box, fitted planes, and dense cloud. Verify transform direction, camera projection and resolution, depth conventions and invalid values, scale, and units.} Preserve common alignment and relative scale; document and consistently apply necessary transformations. Use metres only when metric scale is established; otherwise use reconstruction units. \textcolor{blue}{Distinguish measurements from inferences, assumptions, and uncertainty. Resolve conflicting observations before choosing parameters; treat uncertain fits and incomplete masks as supporting evidence, not ground truth.}

Reproduce the building's form, roof, facade, openings, architectural details, dimensions, proportions, relative placement, and construction relationships. Infer unseen portions plausibly and label them. Exclude vegetation, landscaping, garden furniture, lamps, outside decorations, and unrelated background objects. \textcolor{blue}{The bbox.txt is an approximation; allow small evidence-supported extensions rather than distort measured geometry to fit rough bounds.}

[R04] Analysis resources
Create and inspect useful overviews and analysis aids as needed. Save them and their reproducing code under \textbf{[}LoC1 \textcolor{blue}{LoC2} LoC3\textbf{]}/agent_analysis, \textcolor{red}{organized by round and separate from model deliverables. Record measurement units and uncertainty.}

[R05] Standalone parametric source
All model creation and modification must originate in agent-authored Python using bpy. Write and update the complete program in \textbf{[}LoC1 \textcolor{blue}{LoC2} \textcolor{red}{LoC3}\textbf{]}/temp_blender_code.py; never substitute manual .blend edits for source changes. The program must reconstruct the entire model from a fresh Blender scene, including modelling functions, meaningful dimension/placement parameters, materials, lighting, and animations. Embed required analysis-derived modelling data and output settings. Do not depend on other modelling scripts, existing .blend files, generated geometry, job files, or mandatory command-line arguments. Original resources and explicitly declared libraries may be used.

[R06] Structure and functioning assemblies
Establish stable names/identifiers for elements, doors, windows, materials, and collections; document any renaming of the files and its correspondence. Organize meaningful, separate elements in an IFC-like hierarchy: Project -> Site -> Building -> Storey -> Element groups -> Individual elements. Connect elements according to plausible construction/mechanical relationships; correct unintended gaps, disconnected parts, and implausible intersections.

Doors and windows must have fixed frames and movable leaves/sashes/panels with appropriate hinges, pivots, or sliding axes. Include opening AND closing animations with plausible motion and no obvious collisions. Label inferred mechanisms. Define and document a closed-state frame with correct dimensions and placement; save the .blend at that frame and use it for comparison renders. Preserve any closed-state interface/metadata explicitly supplied by the workflow; if none is supplied, document how to select the closed-state frame.

[R07] Comparison rendering
Use the reference camera poses, projection settings, and native resolution for comparison renders, with doors and windows in the closed state defined in \textbf{[}R05 \textcolor{red}{R06}\textbf{]}. \textcolor{blue}{Keep camera settings unchanged between revisions for before/after comparisons.}

[R08] Validation and uncertainty
Check the model against all available geometric and visual evidence and correct supported errors. \textcolor{red}{Treat quantitative metrics as supporting evidence, not ground truth. Compare depth only after matching conventions and units, and only where both depths are finite and positive inside the reference house mask. Skip views with no valid reference depth or overlap.}

\textcolor{red}{Account for occlusion, inaccurate/incomplete masks, uncertain depth, and inferred structures. Do not remove plausible geometry solely because observations are missing or mistake unknown lighting for geometric errors. Distinguish intended improvements from those verified by inspection or evaluation.}

[R09] Reproducibility
Save Python source for every custom tool and intermediate result, including measurements, fits, overviews, overlays, visualizations, metrics, and models. Keep sources with their outputs and document inputs/metadata, execution commands, library versions, parameters/thresholds, random seeds, unit/coordinate conversions, material assumptions, and inferred geometry sufficiently to reproduce the work.

[R10] Deliverables and round records
\textcolor{red}{Execute the modelling program for initialization and every refinement round. Preserve each round in a separate LoC3 subdirectory containing:}
- A complete executable temp_blender_code.py and a versioned .txt copy with identical contents.
- The .blend generated by executing that program.
- Model/comparison renders and useful diagnostic outputs.
- A record of evidence and measurements used, inferred features, changes, validation results, and every identified discrepancy: supporting views, corrections, and unresolved issues with concrete reasons.
\textcolor{red}{Retain all analysis outputs and reproducing sources under R04/R09, with their locations recorded in the round record. Keep LoC3/temp_blender_code.py updated to the latest complete program, and preserve earlier round snapshots. End each round with a brief report of results, artifact locations, and unresolved questions, distinguishing measured or visually supported improvements from unverified assumptions.}
\textbf{[}
# Start of modelling

Please start modelling following the established context and contract.

\textcolor{red}{# Initialization}

\textcolor{red}{Begin the initial model under the established Context and Contract. Follow this workflow in order:}

\textcolor{red}{1. Measure: Fit surfaces and measure dimensions, proportions, and element positions to the finest detail reliably supported by the data. Record units, uncertainty, and inferences.}
\textcolor{red}{2. Verify: Project fitted boundaries and labelled measurements onto reference images. Inspect across views and resolve conflicts before modelling.}
\textcolor{red}{3. Build: Use the measurements together with all available evidence to create the complete parametric Blender model, from major structures to architectural details and functional assemblies.}
\textcolor{red}{4. Validate: Execute, compare against the evidence, and correct supported errors.}
\textcolor{red}{5. Deliver: Save the artifacts required by the Contract and briefly report how measurements informed the model, inferred features, and unresolved issues.}


\textcolor{red}{# Refinement}

\textcolor{red}{Refine the existing model under the established Context and Contract.}

\textcolor{red}{Perform 5 rounds, selecting 5 new, clear, complementary reference views per round to investigate different perspectives and unresolved issues. Do not repeat selected views across those rounds. If fewer than 5 unused suitable views remain, use those available and report the limitation. Complete all 5 rounds unless the evidence prevents further meaningful work; explain any early stop.}

\textcolor{red}{For each round:}
\textcolor{red}{1. Render the current model from the selected reference cameras under Contract R07. Prepare and inspect:}
   \textcolor{red}{- Reference photographs and model renders side by side.}
   \textcolor{red}{- Pixel-aligned reference/render overlays with adjustable transparency.}
   \textcolor{red}{- Translucent cyan model and aligned orange point cloud overlaid from the same camera, in the same coordinate system and scale.}
   \textcolor{red}{- Reference/rendered depth comparisons, depth-error maps, and useful crops.}

\textcolor{red}{2. Inspect every full view before crops, in this order: geometry soundness (gaps, intersections, disconnected/distorted parts, connections); major structure (volumes, roof, facade planes, extensions, orientation, placement, scale, proportions); then details (openings, frames, doors/windows, and smaller elements). Cross-check structural changes across selected views.}

\textcolor{red}{3. Correct every actionable, evidence-supported discrepancy. Rebuild incorrect assemblies when needed and adjust connected elements consistently. Preserve verified features, but do not retain demonstrably incorrect geometry or defer supported corrections merely to be conservative.}

\textcolor{red}{4. Evaluate depth error, silhouette IoU between the projected model mask and reference house mask, and point-cloud-to-model distance under Contract R08. Rerender from the same cameras, and inspect whether fixes resolved the discrepancies or introduced new ones. Make further supported corrections before completing the round. Prioritize structure over cosmetics.}

\textcolor{red}{5. Save the complete round result and report under Contract R09-R10 before continuing to the next round.}
\textbf{]}
\end{Verbatim}

\subsection{SAM3}\label{subsec:SAM3_prompts}

In the following we describe how we used SAM3 to compute the binary segmentation masks for the \textbf{2D IoU} computation. We also describe how we computed segmentation masks for objects unrelated to the house in each photograph and used these masks to superimpose the selected pixels onto the corresponding rendered image.

\subsubsection{2D IoU Computation}

We use SAM3 to compute the binary segmentation masks of the dataset photographs and the superimposed rendered images, so we can compare them using the \textbf{2D IoU} metric. For each \{photograph, image\}-pair we use the following 11 prompts: "window", "door", "roof", "wall", "chimney", "garage", "balcony", "terrace", "column", "beam",
"stair", "building", "house", "porch", "canopy", "railing" to compute the masks. If there are multiple instances of an object e.g. "window" in a photograph or image, we combine the binary segmentation masks of all the segmented instances by taking the union of the masks. Consequently, there is only one binary segmentation mask per "prompt", if SAM3 detected an instance of that object. We then compute the \textbf{2D IoU} per mask. We report the averaged result over all masks and all \{photograph, image\}-pair.

\subsubsection{Segmentation masks of objects unrelated to the house}

We also used SAM3 to compute the binary segmentation masks of the objects unrelated to the house for each photograph. The prompts for generating the segmentation masks are as follows: "vegetation", "trees", "bushes", "grass", "flowers", "gardens", "fences", "outdoor furniture", "vehicles", "pathways", "objects surrounding the house", "Segment all objects and elements in the outdoor environment that are not part of the house, including vegetation, trees, bushes, grass, flowers, gardens, fences, outdoor furniture, vehicles, pathways, and other surrounding objects". Once every segmentation mask for a photograph was obtained, we combined them all by taking the union of them to obtain a single segmentation mask. We then superimposed the photograph’s pixels selected by the combined segmentation mask onto the corresponding rendered image.

\section{3D Building Model visualization}\label{sec:supp_3D_building_visualizaitons}

\begin{figure}[t]
    \centering
    \includegraphics[width=0.9\textwidth]{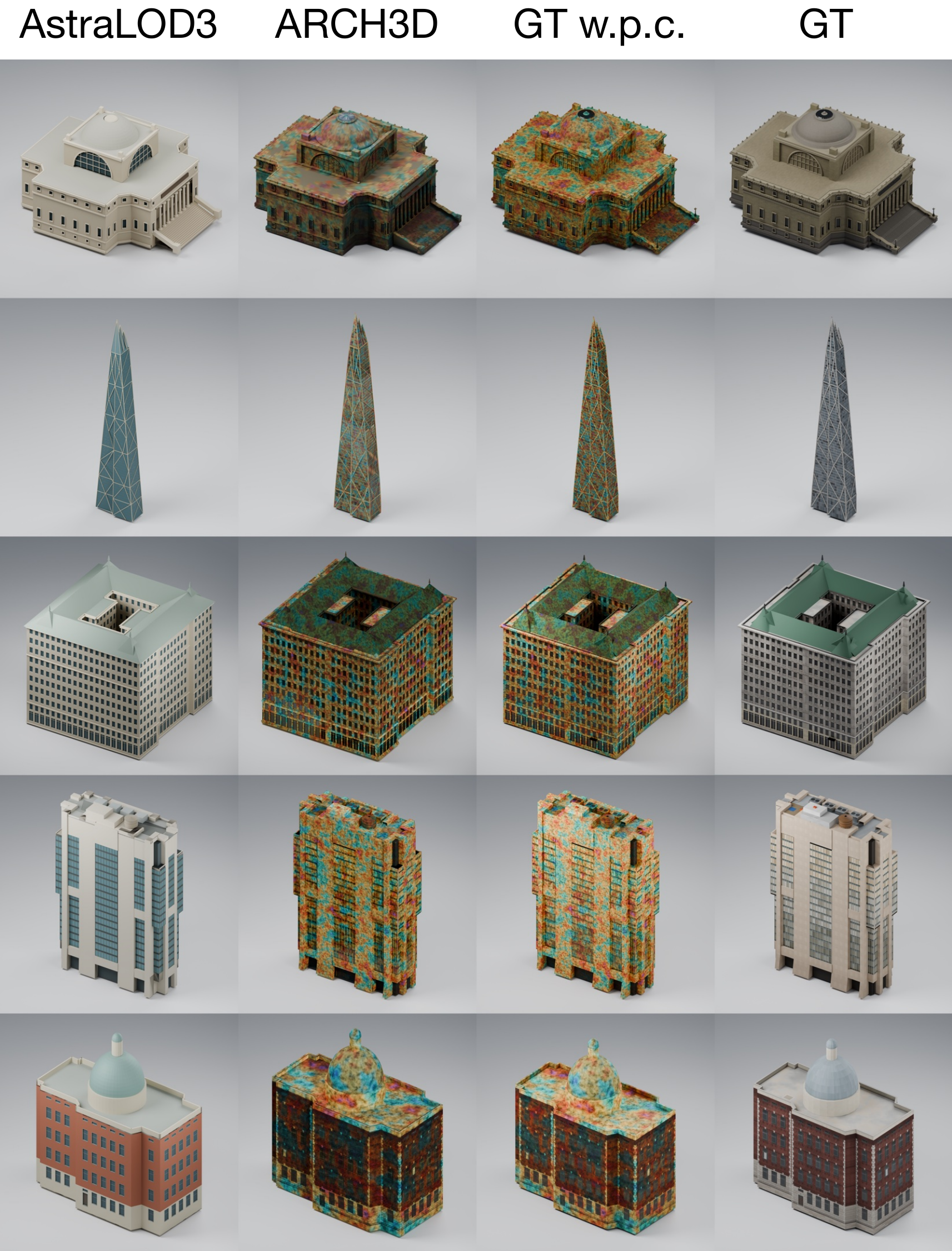}
     \caption{A gallery of the first five building models from the \textit{NYC} dataset made with AstraLOD3 \cite{pantoja-rosero2026a} and our method, ARCH3D, and illustrated next to the ground truth (\textit{GT}) 3D synthetic building models and the same 3D synthetic buildings with procedural colours (\textit{GT w.p.c.}). Note the overlay of texture on the synthetic building models, which was done, as COLMAP failed to make a dense reconstruction due to the repetitive structures on the synthetic 3D buildings}
\label{fig:Gallery_NYC_supplementary_figure_1}
\end{figure}

\begin{figure}[t]
    \centering
    \includegraphics[width=0.9\textwidth]{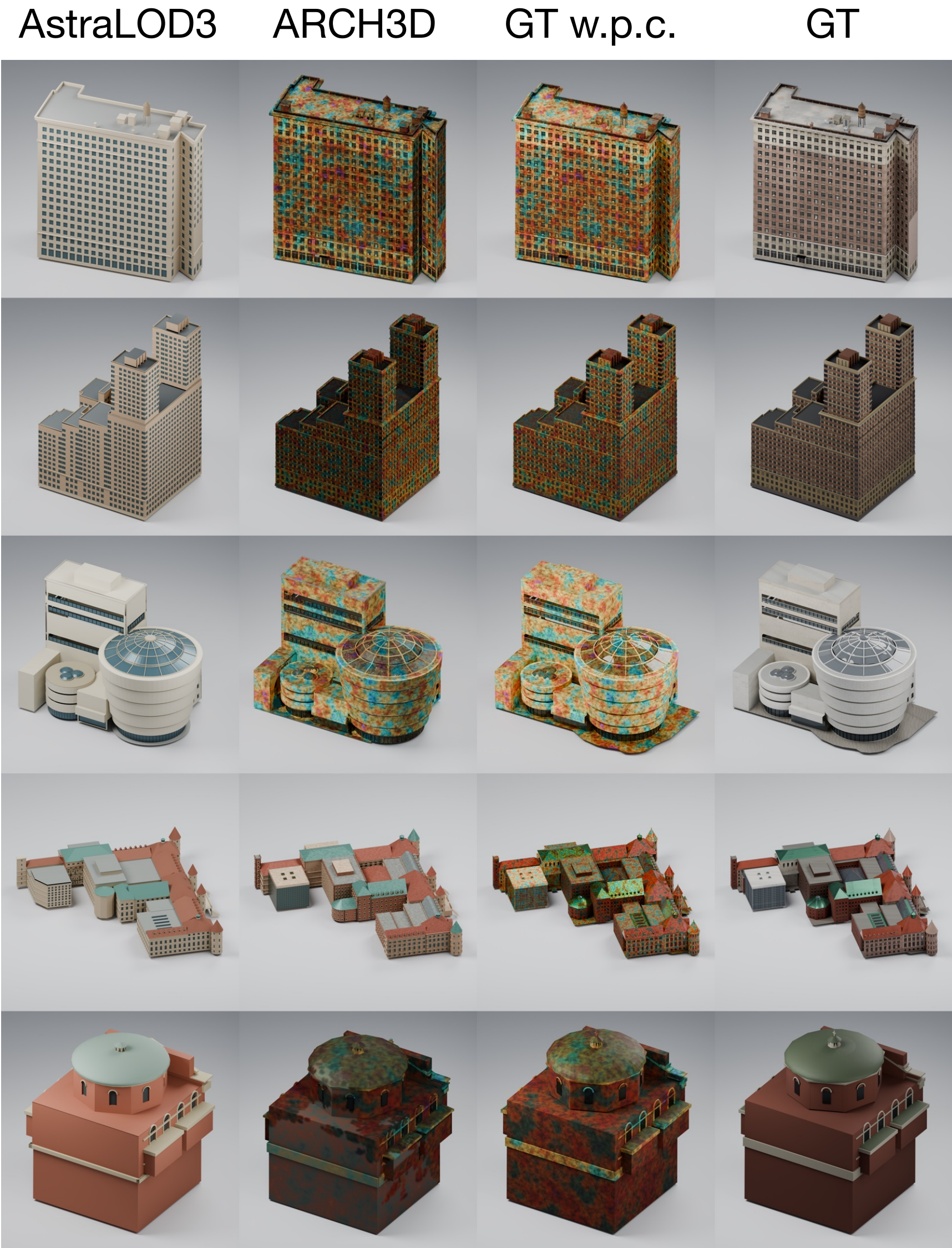}
     \caption{A gallery of the last five building models from the \textit{NYC} dataset made with AstraLOD3 \cite{pantoja-rosero2026a} and our method, ARCH3D, and illustrated next to the ground truth (\textit{GT}) 3D synthetic building models and the same 3D synthetic buildings with procedural colours (\textit{GT w.p.c.}). Note the overlay of texture on the synthetic building models, which was done, as COLMAP failed to make a dense reconstruction due to the repetitive structures on the synthetic 3D buildings}
\label{fig:Gallery_NYC_supplementary_figure_2}
\end{figure}

In \cref{fig:Gallery_NYC_supplementary_figure_1} and \cref{fig:Gallery_NYC_supplementary_figure_2} we display qualitative results of 3D building models generated by both ARCH3D and AstraLOD3 based on synthetic buildings from the \textit{NYC} dataset shown side by side. We manually adjusted the lighting for each building model to facilitate visual evaluation of its geometry and appearance. Note that the repetitive structure on the synthetic buildings made dense reconstruction from rendered images difficult for COLMAP. We therefore applied new textures to the buildings and used the same textures when rendering the reference images, ensuring consistent appearance between the images and the coloured point clouds produced by COLMAP. ARCH3D does not inherently depend on COLMAP, and alternative methods such as VGGT \cite{wang2025vggt} could be used. We did however retain COLMAP for consistency across experiments. From the figures it is seen that ARCH3D reproduces this overlaid texture, whereas AstraLOD3 resorts to making a more realistic-looking texture.

\begin{figure}[t]
    \centering
    \includegraphics[width=0.9\textwidth]{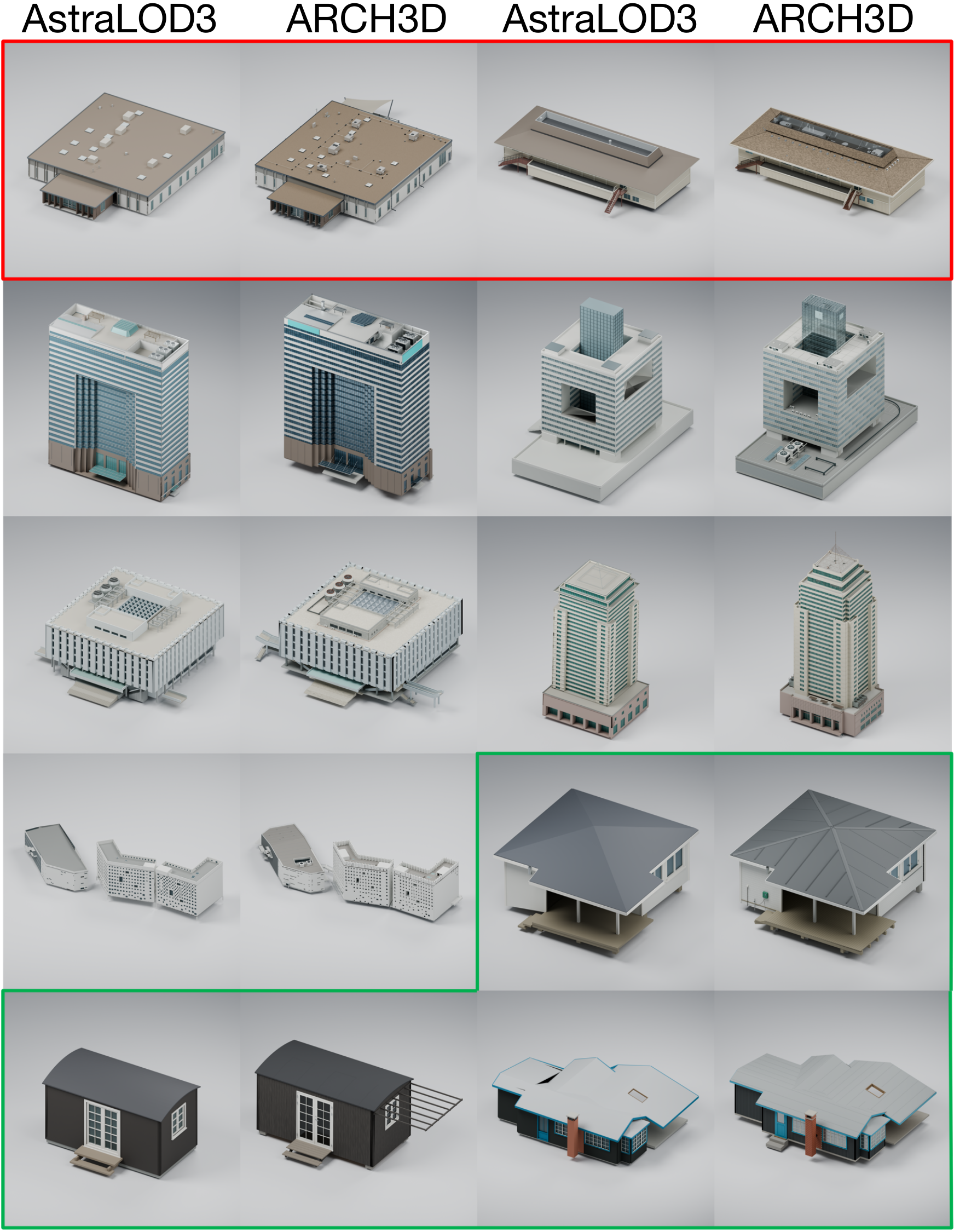}
     \caption{A gallery of selected models from the three datasets (\textit{Matt1up} (highlighted in red), \textit{TwinTex} and \textit{Cottages} (highlighted in green)) made with AstraLOD3 \cite{pantoja-rosero2026a} and our method, ARCH3D. Notice that the lighting is the same for all generated 3D building models.}
\label{fig:Gallery_real_supplementary_figure}
\end{figure}

Following the same principle, we visualize all models reconstructed from real building datasets (\textit{Cottages, Matt1up, and TwinTex}) by ARCH3D and AstraLOD3 in \cref{fig:Gallery_real_supplementary_figure}, again with lighting adjusted for each model to facilitate comparison of geometry and appearance.

\begin{figure}[]
    \centering
    \includegraphics[width=0.9\textwidth]{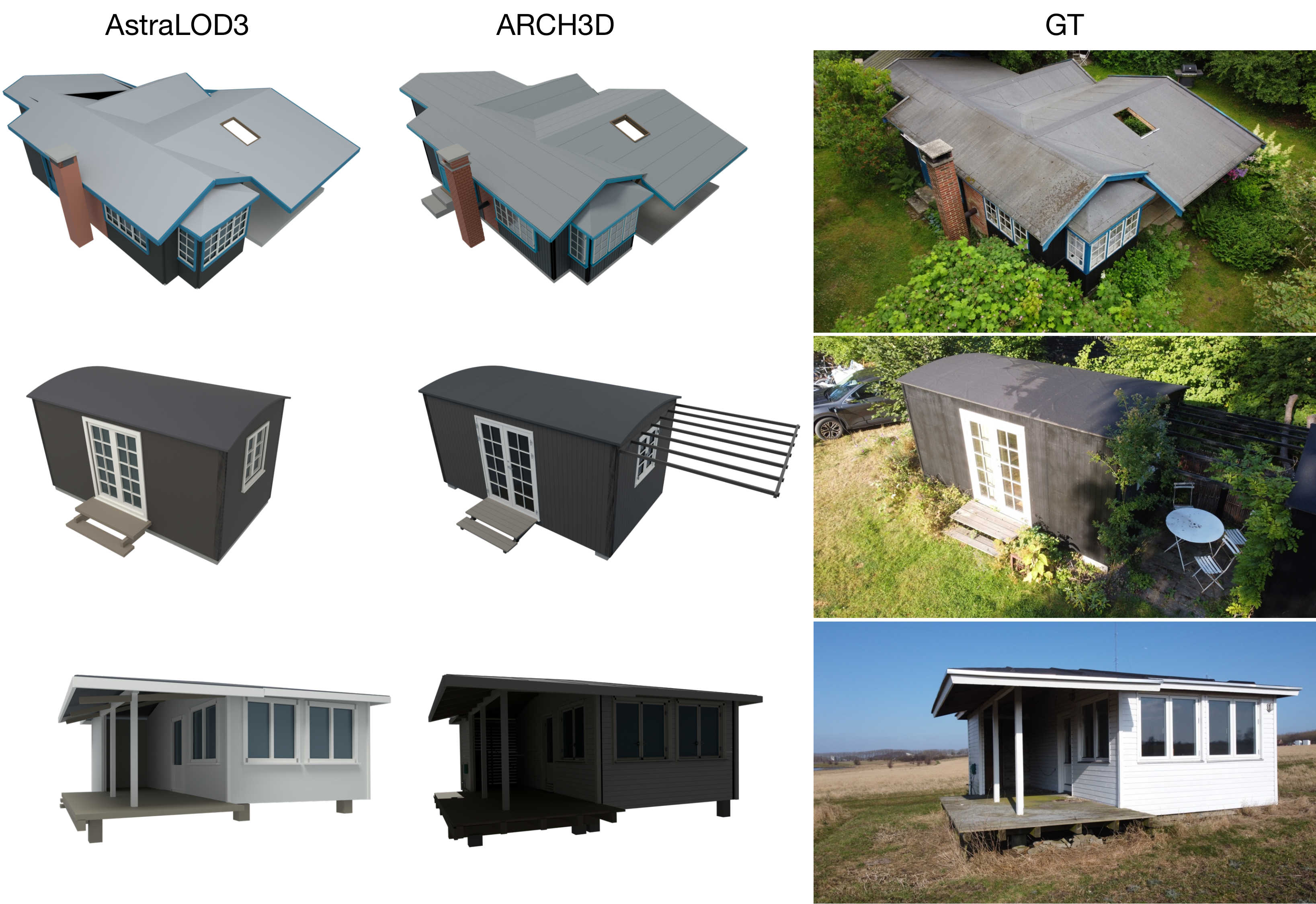}
     \caption{A gallery of the 3D building models from the \textit{Cottages} dataset created by AstraLOD3 and our method, ARCH3D, rendered from the same view point as the photograph (GT) from the dataset}
\label{fig:Gallery_Cottages_next_to_GT_photograph}
\end{figure}

\begin{figure}[]
    \centering
    \includegraphics[width=0.9\textwidth]{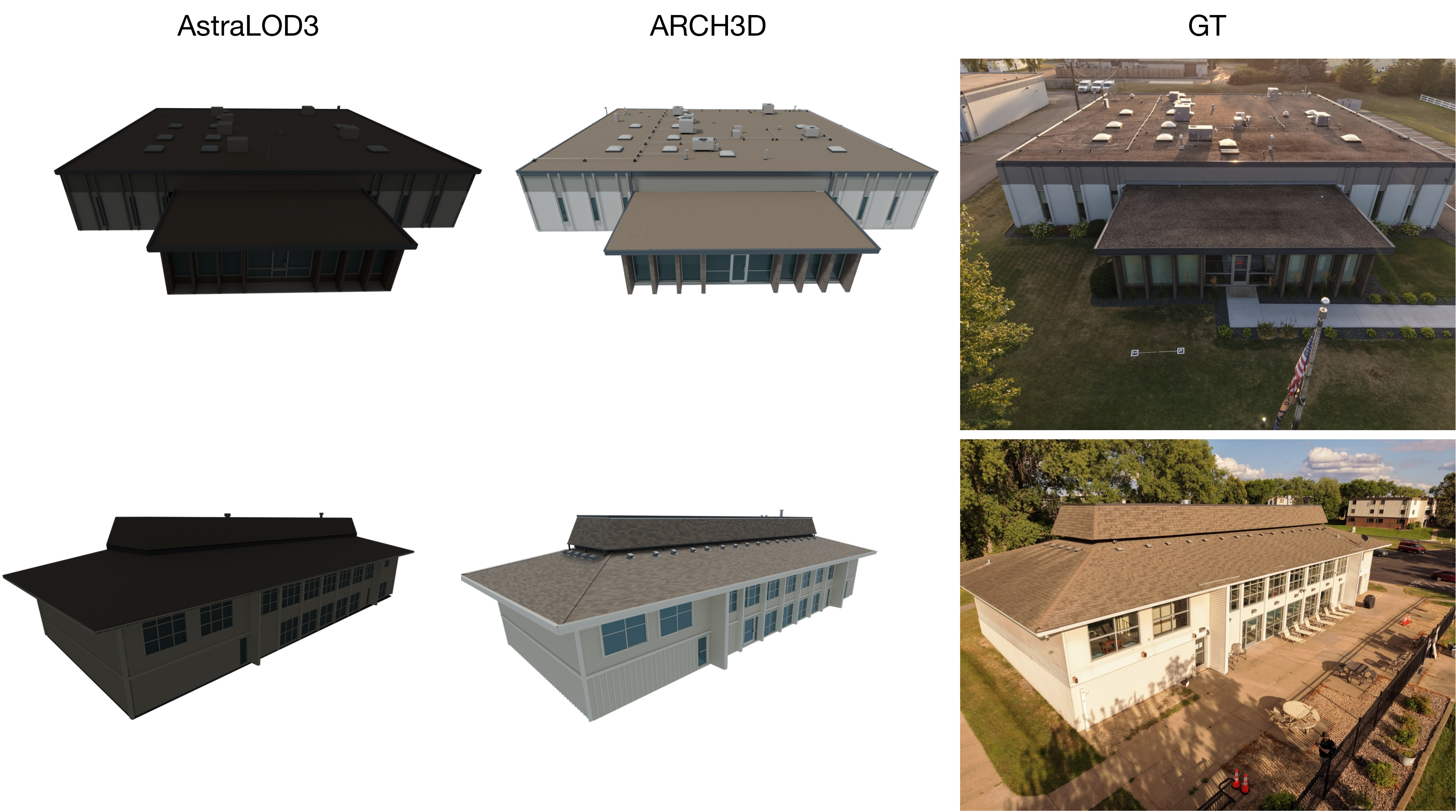}
     \caption{A gallery of the 3D building models from the \textit{Matt1up} dataset created by AstraLOD3 and our method, ARCH3D, rendered from the same view point as the photograph (GT) from the dataset}
\label{fig:Gallery_Matt1up_next_to_GT_photograph}
\end{figure}

\begin{figure}[]
    \centering
    \includegraphics[width=0.9\textwidth]{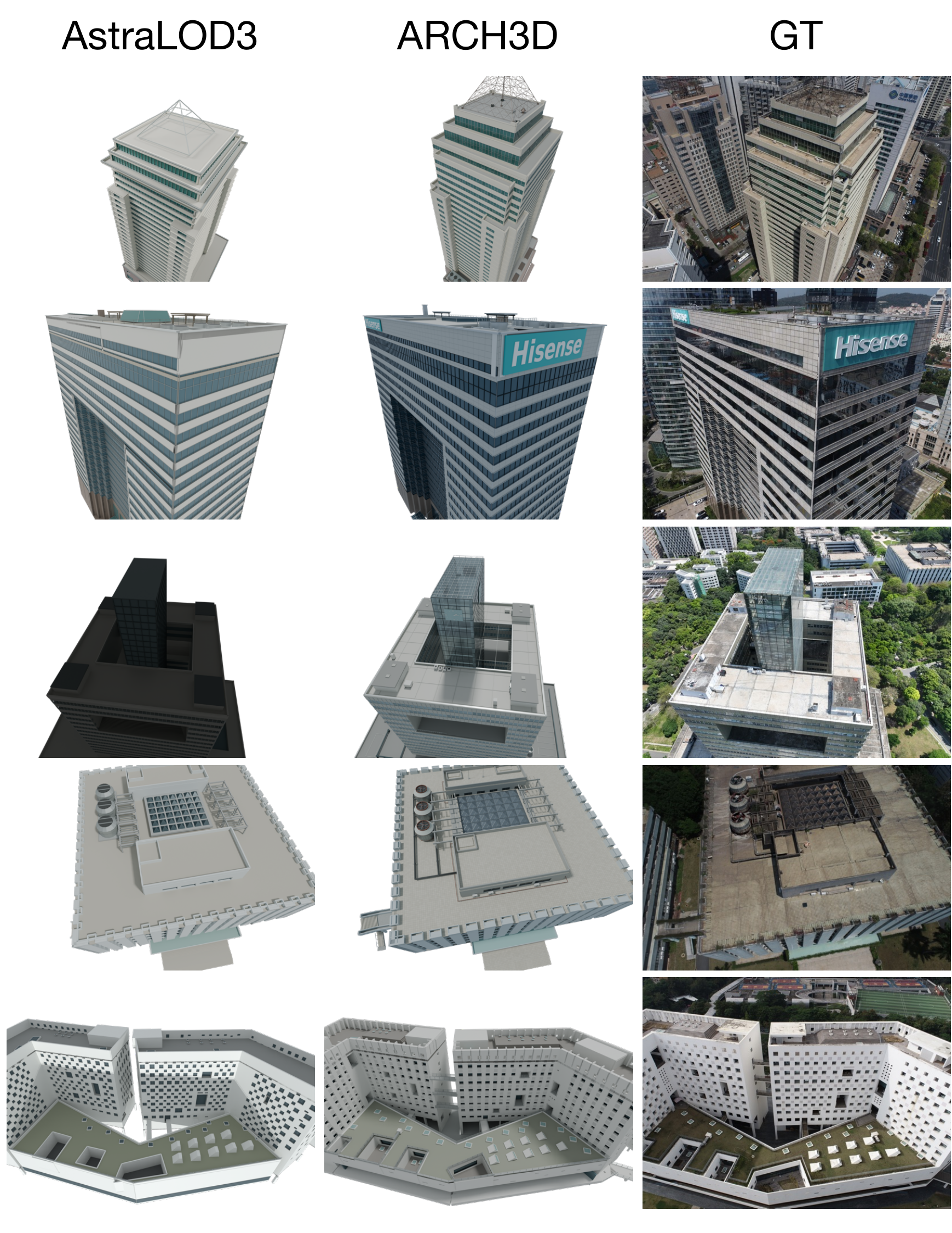}
     \caption{A gallery of the 3D building models from the \textit{TwinTex} dataset created by AstraLOD3 and our method, ARCH3D, rendered from the same view point as the photograph (GT) from the dataset}
\label{fig:Gallery_TwinTex_next_to_GT_photograph}
\end{figure}

\begin{figure}[]
    \centering
    \includegraphics[width=\textwidth]{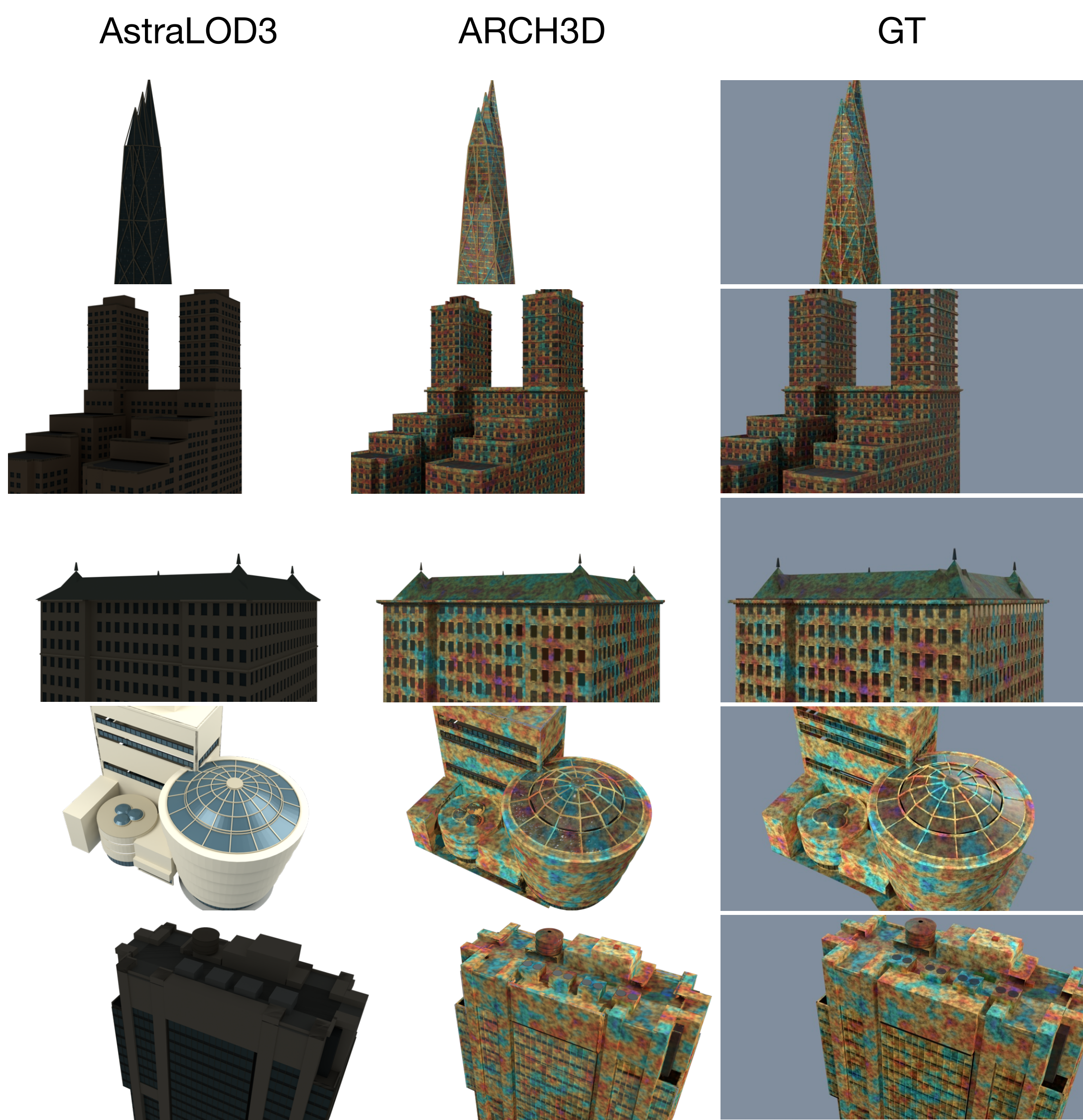}
     \caption{A gallery of the first five 3D building models from the \textit{NYC} dataset created by AstraLOD3 and our method, ARCH3D, rendered from the same view point as the reference image (GT) from the dataset}
\label{fig:Gallery_NYC1_next_to_GT_photograph}
\end{figure}

\begin{figure}[]
    \centering
    \includegraphics[width=\textwidth]{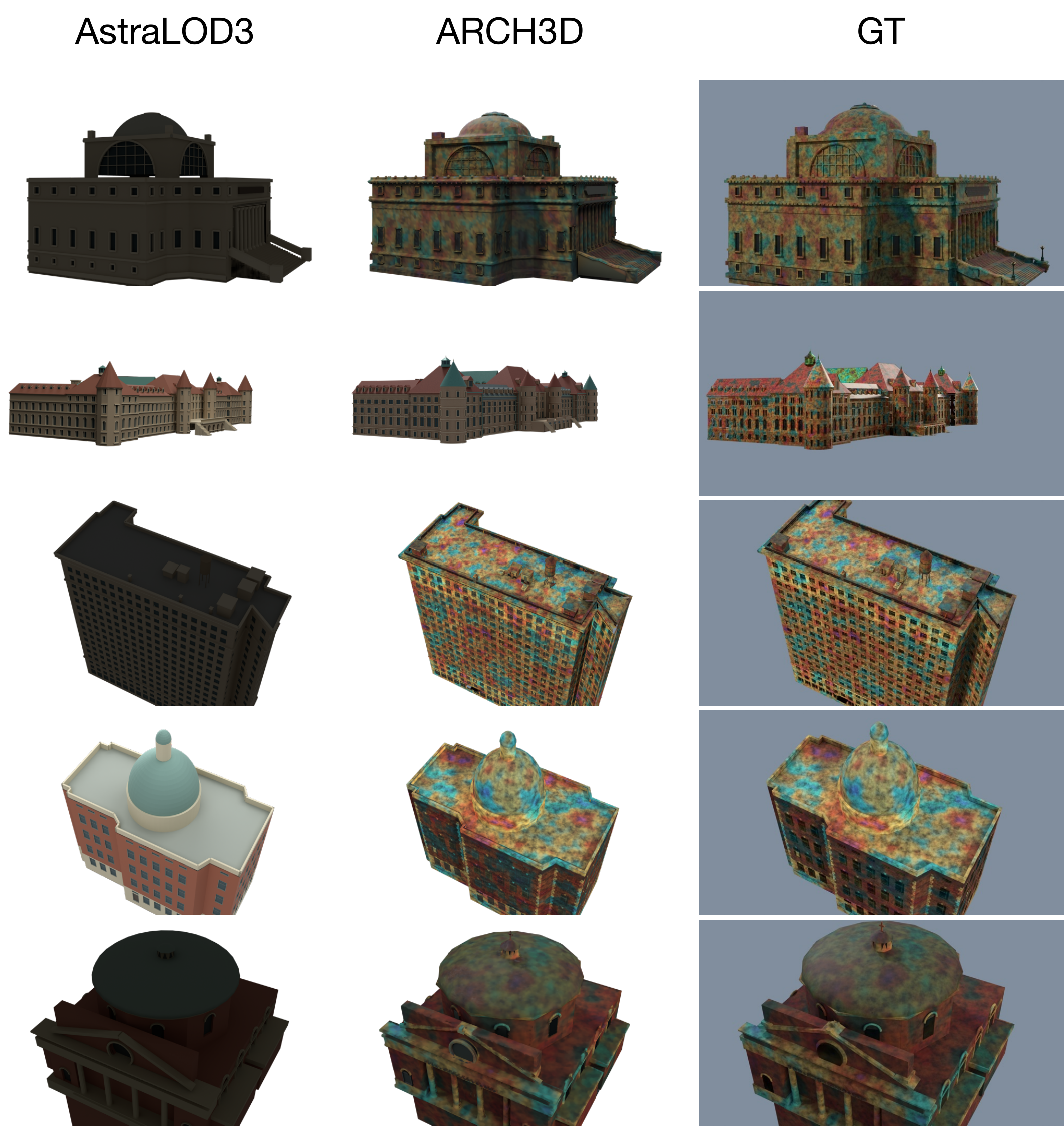}
     \caption{A gallery of the last five 3D building models from the \textit{NYC} dataset created by AstraLOD3 and our method, ARCH3D, rendered from the same view point as the reference image (GT) from the dataset}
\label{fig:Gallery_NYC2_next_to_GT_photograph}
\end{figure}

In \cref{fig:Gallery_Cottages_next_to_GT_photograph}, \cref{fig:Gallery_Matt1up_next_to_GT_photograph}, \cref{fig:Gallery_TwinTex_next_to_GT_photograph}, \cref{fig:Gallery_NYC1_next_to_GT_photograph}, \cref{fig:Gallery_NYC2_next_to_GT_photograph} we show additional qualitative results, where we compare renderings of the 3D building models generated by AstraLOD3 and ARCH3D to the ground truth (GT) photographs or reference images across all datasets. It is clearly seen that ARCH3D manages to capture both the overall structure and the small details of the 3D buildings better than AstraLOD3.

\section{Additional evaluations}\label{sec:supp_additional_evaluations}

In the following we report the results of our additional experiments.

\subsection{LoC-3+}\label{subsec:supp_LoC3_add_introduction}
We conduct an experiment on the cottage depicted in figure \ref{fig:teaser} with another level of control called LoC-3+ and report the results (\textbf{2D IoU} and \textbf{pc-Dist.} introduced in the main manuscript) over 5 trials in \cref{tab:New_models_and_LOC_results} in \cref{subsec:New_GPT_models}. Compared with LoC-3, LoC-3+ further tightens the harness through algorithmic control of the refinement pipeline, restricting the agent’s role to analysing the supplied evidence and constructing or revising building models. It is seen how LoC-3+ performs slightly better than LoC-3, but this comes at the expense of longer computation time and higher token usage, as seen in the plot in \cref{fig:new_computational_consumption}. Pseudocode of the algorithmic control in LoC-3+ is available in \cref{sec:Loc3+code}.\\

The prompts for LoC-3+ largely follow those of LoC-3, with additional requirements in \textit{Contract} and a revised \textit{Refine} prompt specifying a single refinement iteration, as detailed below:
\begin{Verbatim}[
    fontsize=\small,
    breaklines=true,
    breakanywhere=true,
    commandchars=\\\{\}
]
## Workflow interface
Read job_file during authoring to obtain job.output_dir and job.blend_file. Embed the requested .blend destination and required settings as script constants; the completed program must run directly with Blender's --python option without reading a job file or requiring runtime arguments. Save a standalone .blend. Use accepted_model_file as an editing reference when supplied; do not execute or import earlier modelling scripts or load their .blend files/generated geometry. The workflow controls round count, view selection, official comparisons, and acceptance. Each Refine request asks for one revision only, not five rounds.

## Model-to-renderer contract
Save the .blend at the requested destination with the intended scene active and: 
- All building geometry as realized mesh objects under ReconstructionBuilding (nested collections allowed), visible in viewport and render, in reference coordinates and units. No other renderable geometry; lights may be outside it.
- Materials, world/lighting settings, and door/window transform animations.
- scene['reconstruction_closed_frame']: an integer frame where all doors/windows are closed, with opening and closing animation keyframes preserved.
- reconstruction_movable=True on each movable door/window leaf or assembly root.
- scene['reconstruction_closed_parts']: a JSON string listing every tagged object and its known closed world transform, using this entry format: {"object":"stable object name","matrix_world":[[4 values],[4 values],[4 values],[4 values]]}. Use [] only when there are no movable parts. At the declared closed frame, object transforms must match this registry.


# Refinement

Refine the current accepted model under the established Context and Contract. Perform one revision for this round only; the workflow controls subsequent rounds. The preceding selection turn chose five unused, complementary views. Comparisons exclude views with no valid reference depth or no valid reference/render overlap.

Exact paths are supplied in TURN UPDATE. Open comparisons_file for the per-view index and depth/render protocols; comparison_views contains the same view entries:

- reference_image, render_image, comparison_images: photographs, renders, and comparison panels at the reference cameras and native resolution.
- point_cloud_overlay: translucent cyan model and aligned orange cloud.
- reference_depth, rendered_depth, depth_error_map, depth_error, worst_region_crops: depth evidence, valid-overlap mean error, and crops with original coordinates.
- silhouette_iou, reference_silhouette, rendered_silhouette: overlap and masks.
- point_cloud_distance, point_cloud_samples, point_cloud_distances_file: mean surface distance, sample count, and per-sample distances.

Null metrics mean unavailable evidence, not zero error. The index also supplies metrics_file. Use last_decision_file for previous feedback when present.

1. Inspect every full view before crops: geometry soundness (gaps, intersections, disconnected/distorted parts, connections); major structure (volumes, roof, facade planes, extensions, orientation, placement, scale, proportions); then details (openings, frames, doors/windows, smaller elements). Cross-check structural changes across selected views. Prepare pixel-aligned photograph/ render overlays with adjustable transparency when useful, preserving their reproducing source under job.output_dir/agent_analysis.

2. Correct every actionable, evidence-supported discrepancy. Rebuild incorrect assemblies when needed and adjust connected elements consistently. Preserve verified features; do not defer supported corrections merely to be conservative.

3. Interpret depth error, silhouette IoU, and cloud distance under Contract R08. Use diagnostic renders from the same cameras to inspect fixes and new issues. Make further supported corrections before returning. Prioritize structure over cosmetics; record unresolved findings with concrete reasons.

4. Use accepted_model_file as an editing reference and write the complete updated standalone program at context.code_file. Read job_file during authoring and embed job.blend_file; do not read job files at model execution time. Preserve the model-to-renderer interface, including closed-state metadata.

5. Save analysis sources and the round record under job.output_dir/agent_analysis following R09-R10. Report findings, corrections, supporting views, and unresolved issues, distinguishing inspected improvements from unverified intentions.

Return after writing this revision. The workflow rebuilds, archives, renders, evaluates, and decides acceptance; official evaluation is not the agent's task.
\end{Verbatim}

\subsection{New GPT models}\label{subsec:New_GPT_models}
\begin{figure}[H]
    \centering
    \includegraphics[width=\textwidth]{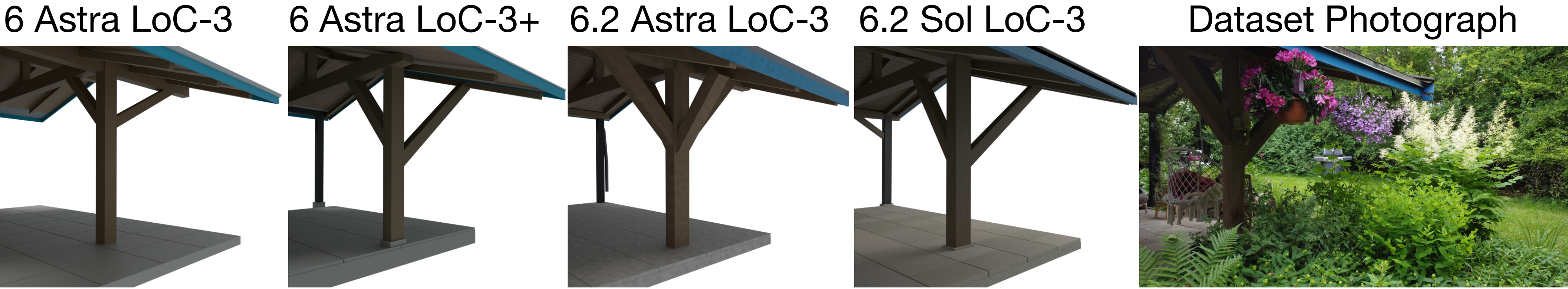}
     \caption{Qualitative results GPT-6 Astra LoC-3 compared to GPT-6 Astra LoC-3+ and the new VLMs, GPT 6.2-Astra and GPT-6.2 Sol. }
\label{fig:qualitative_results_new_vlms}
\end{figure}

After the paper submission deadline, OpenAI released GPT-6.2 Astra and GPT-6.2 Sol, which are improved versions of the model that we used to conduct our experiments. We therefore also run 5 trials for each model (with reasoning effort Medium and LoC-3) and evaluate these using the same metrics \textbf{2D IoU} and \textbf{pc-Dist.}. We report the results in \cref{tab:New_models_and_LOC_results}. It is clearly seen that the performance of the 6.2 models is better than the version 6 models. Furthermore, it is also seen that the performance of GPT-6.2 Astra is very comparable to GPT-6.2 Sol. We also provide qualitative results in \cref{fig:qualitative_results_new_vlms}, where it is seen that all configurations except for LoC-3 (ARCH3D) capture the column in the very left of the photograph. All models exhibit errors in reconstructing the supporting structure on the nearest column, with GPT-6.2 Astra producing the closest match to the reference.

\begin{figure}[H]
    \centering
    \includegraphics[width=\textwidth]{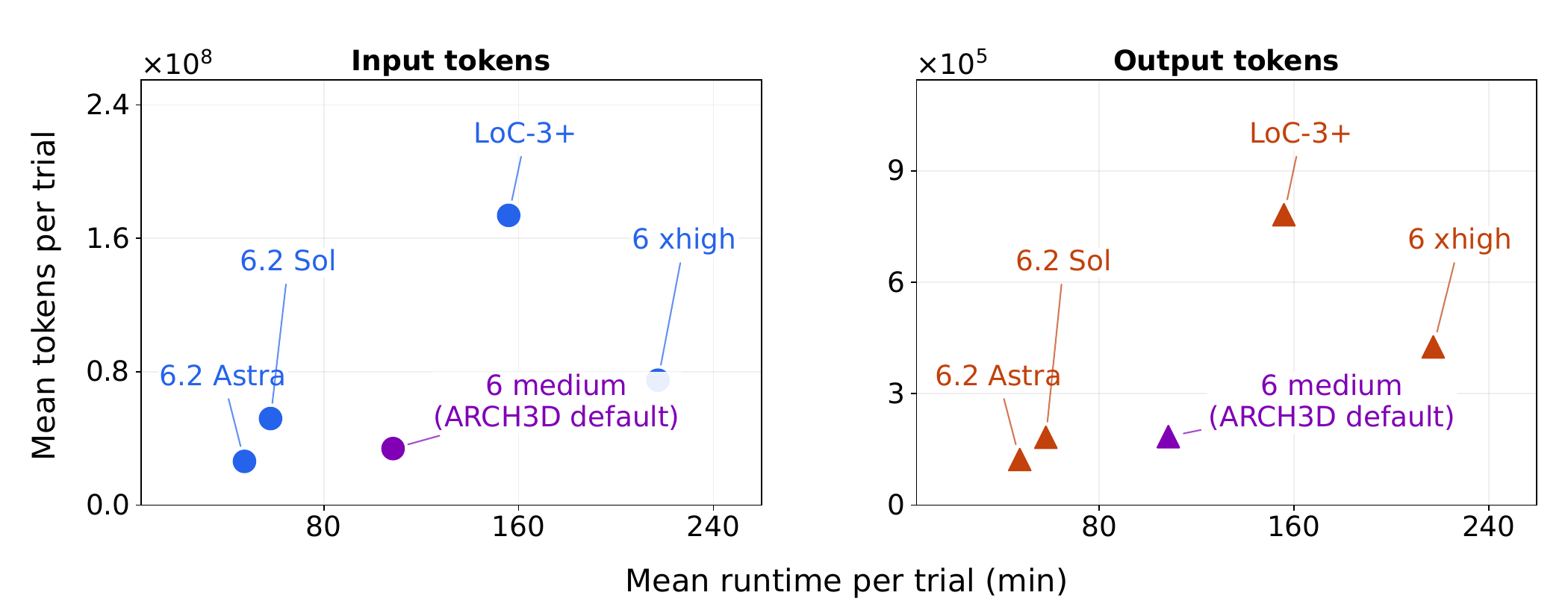}
     \caption{Average execution time and token consumption per trial on the same building in \textit{Cottages} dataset for ARCH3D using models 6.2 Astra and 6.2 Sol, compared with the default configuration (model 6, medium reasoning effort; purple).}
\label{fig:new_computational_consumption}
\end{figure}

The scatter plots in \cref{fig:new_computational_consumption} compare the computational costs of the newly released models. Overall, GPT-6.2 Astra and GPT-6.2 Sol achieve higher reconstruction quality than GPT-6 Astra (cf. \cref{tab:New_models_and_LOC_results}) with comparable token consumption and approximately half the execution time.

\begin{table}[H]
  \centering
  \setlength{\tabcolsep}{10pt}
  \caption{Evaluation of LoC-3+ and the two new models: \textit{"GPT-6 Astra / LoC-3+"}, \textit{"GPT-6.2 Sol / LoC-3"}  and \textit{"GPT-6.2 Astra / LoC-3"}. We compare them against \textit{GPT-6 Astra / LoC-3}, which is default ARCH3D configuration introduced in the main manuscript.}
  \begin{tabular}{@{}lcc@{} }
    \toprule
    \textbf{Method} & \textbf{2D IoU} [$\mu,P_{25},P_{50},P_{75}$] $\uparrow$  & \textbf{pc-Dist} [$\mu,P_{25},P_{50},P_{75}$] $\cdot 10^3$ $\downarrow$  \\
    \midrule  
    \multicolumn{3}{@{}l@{}}{Reasoning effort: Medium for all models}\\
    \midrule
    GPT-6 Astra / LoC-3  & ${0.606}/{0.304}/{0.766}/{0.900}$ & ${8.80}/{1.61}/{4.30}/{9.82}$  \\
    GPT-6 Astra / LoC-3+  & $0.6210/0.350/0.780/0.904$ & $7.75/1.65/4.25/9.12$  \\
    GPT-6.2 Astra / LoC-3  & $0.659/0.476/\mathbf{0.814}/\mathbf{0.919}$ & $5.63/\mathbf{1.02}/2.64/\mathbf{5.74}$  \\
    GPT-6.2 Sol / LoC-3  & $\mathbf{0.662}/\mathbf{0.500}/0.812/0.915$ & $\mathbf{5.55}/1.06/\mathbf{2.56}/5.76$  \\
    \bottomrule
  \end{tabular}
  \label{tab:New_models_and_LOC_results}
\end{table}

\begin{figure*}[b]
    \centering
    \includegraphics[width=\textwidth]{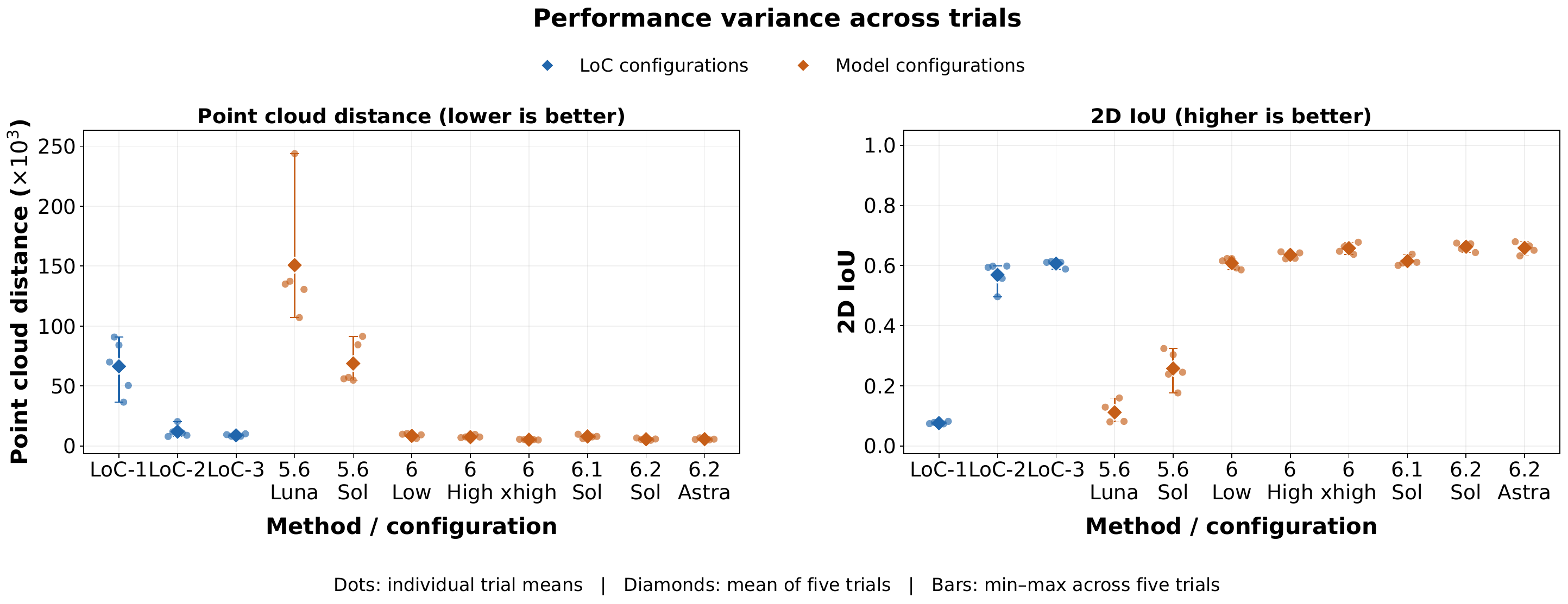}
     \caption{Performance variability across the five repeated trials for each configuration reported in the main manuscript. Point cloud distance ($\times10^3$, lower is better) and \textbf{2D IoU} (higher is better) are shown side by side. Dots represent individual trial means, diamonds indicate the average across trials, and whiskers show the minimum–maximum range.}
\label{fig:variance}
\end{figure*}

\subsection{Performance variability across trials}
We further performed evaluations to assess the performance and consistency of different harness configurations and models, using data from five independent runs (trials) for each configuration. \cref{fig:variance} shows the mean and min–max range of point cloud distance (\textbf{pc-Dist.}) and \textbf{2D IoU} across these runs. Overall, a tighter harness improves both performance and consistency: moving from LoC-1 to LoC-3 reduces point cloud distance, increases \textbf{2D IoU}, and generally narrows the variation across runs. This trend is particularly evident for point cloud distance. Although LoC-1 exhibits a narrow \textbf{2D IoU} range, its consistently low scores indicate consistently poor reconstruction quality. Model choice further affects robustness: the 5.6 Luna shows poorer average performance and greater variability, whereas newer configurations generally achieve better scores with tighter ranges. The strongest configurations perform similarly, suggesting that further improvements from model choice may be modest under a tightly constrained harness. 

\newpage

\twocolumn
\section{Pseudocode for semantic-aware geometry extraction}\label{sec:supp_pseudocode_for_semantic_aware_geometry}

\cref{alg:data_preparation} presents a pseudo-code for semantic-aware geometry extraction from images. $\mathcal{R}_{\mathrm{sp}}$ is the sparse reconstruction results. The set $\mathcal{V}_k$ contains depth-consistent views of point $\mathbf{p}_k$, and $n^+_{kj}$ counts positive mask observations for semantic prompt $j$. The projection $\pi_i$ maps a point to its pixel in image $i$. $\mathbf{1[\cdot]}$ sets the label to 1 if $\cdot$ is true. Camera intrinsics are denoted by $\mathbf{K}_i$, and $\mathbf{T}_{\mathrm{cw},i}$ maps world coordinates to camera coordinates. Point transformations use homogeneous coordinates. Filtering and denoising preserve the correspondence between points $\mathbf{P}$ and labels $\mathbf{L}$.

\begin{algorithm}[t]
\caption{Semantic geometry extraction}
\label{alg:data_preparation}
\begin{algorithmic}[1]
\Require Images $\mathcal{I}$, semantic prompts $\mathcal{S}$
\Require Thresholds $n_{\min}$, $r_{\min}$, and $\epsilon_d$
\Ensure Aligned camera poses $\mathcal{C}_a$, semantically labelled point cloud $\mathbf{P}_a$, depth maps $\mathcal{D}$, semantic masks (instance-level) $\mathcal{M}$, normal maps $\mathcal{N}$

\Statex \textbf{Reconstruction}
\State $(\mathcal{R}_{\mathrm{sp}},\mathcal{C})
       \gets \Call{SfM}{\mathcal{I}}$
\State $(\mathcal{I}_u,\mathcal{C}_u)
       \gets \Call{Undistort}
       {\mathcal{I},\mathcal{R}_{\mathrm{sp}},\mathcal{C}}$
\State $(\mathcal{D},\mathcal{N})
       \gets \Call{MVS}{\mathcal{I}_u,\mathcal{C}_u}$

\Statex \textbf{Semantic filtering}
\State $\mathcal{M}
       \gets \Call{SAM}{\mathcal{I}_u,\mathcal{S}}$
\State $M_{\cup,i} \gets \bigvee_j M_{ij}$
       for each image $i$, union semantic prompt $j$
\State $\mathbf{P}
       \gets \Call{Fuse}
       {\mathcal{D},\mathcal{N},\mathcal{C}_u,\mathcal{M}_{\cup}}$
\For{each point $\mathbf{p}_k$ in $\mathbf{P}$}
    \State $\mathcal{V}_k
           \gets \Call{Visible}
           {\mathbf{p}_k,\mathcal{C}_u,\mathcal{D},\epsilon_d}$
    \For{each semantic prompt $j$}
        \State $n^+_{kj}
               \gets \sum_{i \in \mathcal{V}_k}
               M_{ij}\!\left(\pi_i(\mathbf{p}_k)\right)$
        \State $L_{kj}
               \gets \mathbf{1}\!
               \left[n^+_{kj} \geq
               \max\{n_{\min},r_{\min}|\mathcal{V}_k|\}\right]$
    \EndFor
\EndFor
\State $(\mathbf{P},\mathbf{L})
       \gets \Call{Filter}{\mathbf{P},\mathbf{L}}$
       \Comment{Keep points with any label}
\State $(\mathbf{P},\mathbf{L})
       \gets \Call{Denoise}{\mathbf{P},\mathbf{L}}$

\Statex \textbf{Alignment}
\State $\mathbf{T}_{\mathrm{align}}
       \gets \Call{Align}{\mathbf{P},\mathcal{C}_u}$
\State $\mathbf{P}_a
       \gets \mathbf{T}_{\mathrm{align}}\mathbf{P}$
\State $\mathcal{C}_a
       \gets \{(\mathbf{K}_i,
       \mathbf{T}_{\mathrm{cw},i}
       \mathbf{T}_{\mathrm{align}}^{-1})\}_i$

\end{algorithmic}
\end{algorithm}

\section{Pseudocode for LoC-3+}\label{sec:supp_pseudocode_for_LoC3_add}
\label{sec:Loc3+code}
\cref{alg:loc3plus} provides pseudocode for the algorithmic control implemented for LoC-3+. Here, $A$ denotes the persistent Codex session, $\mathcal{D}$ is the evidence from the semantic-aware geometry extraction, and $P_c$, $P_k$, $P_i$, and $P_r$ denote the context, contract, initialization, and refinement prompts, respectively. $\mathcal{C}$ is a shortlist of unused, spatially diverse reference views, from which the agent selects $V$, a set of views. The evaluation returns metrics $E$ (point-cloud distance and depth errors) and the subset $S\subseteq V$ with valid depth comparisons. $X$ contains diagnostic evidence, including image comparisons, point-cloud overlays, and depth-error crops. Primed symbols denote the candidate reconstruction and its evaluation. The acceptance flag $a$ is determined by the mean depth-error reduction $\Delta$, where $e_v$ and $e'_v$ are the current and candidate models' depth errors in view $v$.

\begin{algorithm}[t]
\caption{LoC-3+ reconstruction}
\label{alg:loc3plus}
\begin{algorithmic}[1]
\Require Evidence $\mathcal{D}$, prompts $P_c,P_k,P_i,P_r$,
         rounds $N_r$, views per round $N_v$
\Ensure Reconstructed blender model $M$
\State $A \gets \Call{Codex}{P_c,P_k}$
\State $M \gets \Call{Build}{A,P_i,\mathcal{D}}$;
       $\mathcal{U} \gets \emptyset$
\For{$r=1,\ldots,N_r$}
    \State $\mathcal{C} \gets
        \Call{Shortlist}{\mathcal{D}\setminus\mathcal{U}}$
    \State $V \gets \Call{SelectValid}{A,\mathcal{C},N_v}$
    \State $\mathcal{U} \gets \mathcal{U}\cup V$
    \State $(E,S) \gets \Call{Evaluate}{M,V}$
    \If{$S=\emptyset$}
        \State \textbf{continue}
    \EndIf
    \State $X \gets \Call{Diagnostics}{\mathcal{D},M,V,E}$
    \State $M' \gets \Call{Build}{A,P_r,M,X}$
    \State $(E',S') \gets \Call{Evaluate}{M',V}$
    \State $a \gets \mathrm{false}$
    \If{$S'=S$ and all depth scores are valid}
        \State $\Delta \gets
            \frac{1}{|S|}\sum_{v\in S}(e_v-e'_v)$
        \State $a \gets (\Delta>10^{-8})$
    \EndIf
    \If{$a$}
        \State $M \gets M'$
    \EndIf
\EndFor
\State \Return $M$
\end{algorithmic}
\end{algorithm}

\section{Data availability}\label{sec:supp_data_availability}

The \textit{TwinTex} \cite{TwinTex23} and \textit{Matt1up} \cite{guertin2026drone} datasets, along with the source data for \textit{NYC} \cite{turkcan2026valdrada}, are publicly available online. Upon acceptance, we will release our modified \textit{NYC} dataset and the photographs of the buildings from the \textit{Cottages} dataset.
{
    \small
    \twocolumn
    \bibliographystyle{ieeenat_fullname}
    \bibliography{main}

@String(ICCV= {Int. Conf. Comput. Vis.})

@String(ECCV= {Eur. Conf. Comput. Vis.})

@String(TOG= {ACM Trans. Graph.})

@String(ICCV  = {ICCV})

@String(ECCV  = {ECCV})

@String(TOG   = {ACM TOG})

@article{mildenhall2020a,
  author = {Mildenhall, Ben and Srinivasan, Pratul P. and Tancik, Matthew and Barron, Jonathan T. and Ramamoorthi, Ravi and Ng, Ren},
  title = {NeRF: Representing Scenes as Neural Radiance Fields for View Synthesis},
  language = {eng},
  format = {article},
  editor = {Vedaldi, A, Bischof, H, Frahm, JM, Brox, T},
  journal = {Lecture Notes in Computer Science (including Subseries Lecture Notes in Artificial Intelligence and Lecture Notes in Bioinformatics)},
  volume = {12346},
  pages = {405-421},
  year = {2020},
  issn = {30049946, 16113349, 03029743},
  isbn = {3030584518, 3030584526, 9783030584511, 9783030584528},
  publisher = {Springer Science and Business Media Deutschland GmbH},
  doi = {10.1007/978-3-030-58452-8_24}
}

@article{barron2023a,
  author = {Barron, Jonathan T. and Mildenhall, Ben and Verbin, Dor and Srinivasan, Pratul P. and Hedman, Peter},
  title = {Zip-NeRF: Anti-Aliased Grid-Based Neural Radiance Fields},
  language = {eng},
  format = {article},
  journal = {Proceedings of the Ieee International Conference on Computer Vision},
  pages = {19640-19648},
  year = {2023},
  issn = {23807504, 15505499},
  isbn = {9798350307184},
  publisher = {Institute of Electrical and Electronics Engineers Inc.},
  doi = {10.1109/ICCV51070.2023.01804}
}

@article{kerbl2023a,
  author = {Kerbl, Bernhard and Kopanas, Georgios and Leimkuehler, Thomas and Drettakis, George},
  title = {3D Gaussian Splatting for Real-Time Radiance Field Rendering},
  language = {eng},
  format = {article},
  journal = {Acm Transactions on Graphics},
  volume = {42},
  number = {4},
  pages = {3592433},
  year = {2023},
  issn = {15577368, 07300301},
  publisher = {Association for Computing Machinery},
  doi = {10.1145/3592433}
}

@article{yin2026a,
  author = {Yin, Shaofeng and Ge, Jiaxin and Wang, Zora Zhiruo and Wang, Chenyang and Li, Xiuyu and Black, Michael J. and Darrell, Trevor and Kanazawa, Angjoo and Feng, Haiwen},
  title = {Vision-as-Inverse-Graphics Agent via Interleaved Multimodal Reasoning},
  language = {eng},
  format = {article},
  journal = {Computer Vision – Eccv 2026},
  pages = {557-575},
  year = {2026},
  issn = {16113349, 03029743},
  isbn = {3032374634, 3032374642, 9783032374639, 9783032374646},
  publisher = {Springer Nature Switzerland},
  doi = {10.1007/978-3-032-37464-6_30}
}

@misc{meshy,
    author       = {{Meshy AI}},
    title        = {{Meshy: AI 3D Model Generator}},
    year         = {2026},
    howpublished          = {\url{https://www.meshy.ai}},
    note         = {Accessed: November 1, 2026}
}

@inproceedings{pan2024glomap,
    author={Pan, Linfei and Barath, Daniel and Pollefeys, Marc and Sch\"{o}nberger, Johannes Lutz},
    title={{Global Structure-from-Motion Revisited}},
    booktitle={European Conference on Computer Vision (ECCV)},
    year={2024},
}

@misc{carion2025sam3segmentconcepts,
      title={{SAM 3: Segment Anything with Concepts}},
      author={Nicolas Carion et. al.},
      year={2025},
      eprint={2511.16719},
      archivePrefix={arXiv},
      primaryClass={cs.CV},
      howpublished={\url{https://arxiv.org/abs/2511.16719}},
}

@software{turkcan2026valdrada,
  author  = {Turkcan, Mehmet Kerem},
  title   = {{Valdrada}: a real-data, real-time digital twin of New York City},
  year    = {2026},
  version = {0.6.2},
  howpublished     = {\url{https://github.com/mkturkcan/valdrada}}
}

@article{doris2025a,
  author = {Doris, Anna C. and Alam, Md Ferdous and Heyrani Nobari, Amin and Ahmed, Faez},
  title = {CAD-CODER: AN OPEN-SOURCE VISION-LANGUAGE MODEL FOR COMPUTER-AIDED DESIGN CODE GENERATION},
  language = {eng},
  format = {article},
  journal = {Proceedings of the Asme Design Engineering Technical Conference},
  volume = {3},
  pages = {V03AT03A031},
  year = {2025},
  isbn = {079188922X, 9780791889220},
  publisher = {American Society of Mechanical Engineers (ASME)},
  doi = {10.1115/DETC2025-169758}
}

@article{zhou2026articraft,
  title     = {Articraft: An Agentic System for Scalable Articulated 3D Asset Generation},
  author    = {Zhou, Matt and Li, Ruining and Lyu, Xiaoyang and Song, Zhaomou and Huang, Zhening and Zheng, Chuanxia and Rupprecht, Christian and Vedaldi, Andrea and Wu, Shangzhe},
  journal   = {arXiv preprint arXiv:2605.15187},
  year      = {2026}
}

@article{li2026a,
  author = {Li, Xirui and Shi, Peng and Dong, Mingwen and Zhang, Sheng and Xu, Zhuoyan and Lee, Dongkyu and Chang, Shuaichen and Xiang, Yi and Pan, Lin and Jiang, Jiarong},
  title = {LEGO-Anything: Coding Agents for 3D Scene Reconstruction},
  language = {eng},
  format = {article},
  year = {2026},
  doi = {10.48550/arXiv.2609.36380}
}

@article{pantoja-rosero2026a,
  author = {Pantoja-Rosero, Bryan G.},
  title = {{AstraLOD3: Zero-shot multimodal agentic reconstruction of LOD3 building models}},
  language = {und},
  format = {article},
  year = {2026}
}

@online{openai2026astra,
  author    = {{OpenAI}},
  title     = {{GPT-6 Astra}: A new generation of intelligence},
  year      = {2026},
  howpublished       = {\url{https://openai.com/index/gpt-6-astra/}},
  urldate   = {2026-11-03}
}

@online{openai2026Codex,
  author    = {{OpenAI}},
  title     = {{OpenAI - Codex}},
  year      = {2025},
  howpublished       = {\url{https://openai.com/index/introducing-codex/}},
  urldate   = {2026-11-06}
}

@article{muller2022a,
  author = {Müller, Thomas and Evans, Alex and Schied, Christoph and Keller, Alexander},
  title = {Instant neural graphics primitives with a multiresolution hash encoding},
  language = {eng},
  format = {article},
  journal = {Acm Transactions on Graphics},
  volume = {41},
  number = {4},
  pages = {102},
  year = {2022},
  issn = {15577368, 07300301},
  publisher = {Association for Computing Machinery},
  doi = {10.1145/3528223.3530127}
}

@article{fridovich-keil2022a,
  author = {Fridovich-Keil, Sara and Yu, Alex and Tancik, Matthew and Chen, Qinhong and Recht, Benjamin and Kanazawa, Angjoo},
  title = {Plenoxels: Radiance Fields without Neural Networks},
  language = {eng},
  format = {article},
  journal = {Proceedings of the Ieee Computer Society Conference on Computer Vision and Pattern Recognition},
  volume = {2022-},
  pages = {5491-5500},
  year = {2022},
  issn = {2332564x, 10636919, 25757075},
  isbn = {1665469463, 9781665469463},
  publisher = {IEEE Computer Society},
  doi = {10.1109/CVPR52688.2022.00542}
}

@article{chen2022a,
  author = {Chen, Anpei and Xu, Zexiang and Geiger, Andreas and Yu, Jingyi and Su, Hao},
  title = {TensoRF: Tensorial Radiance Fields},
  language = {eng},
  format = {article},
  editor = {Brostow, G, Farinella, GM, Cisse, M, Avidan, S, Hassner, T},
  journal = {Lecture Notes in Computer Science (including Subseries Lecture Notes in Artificial Intelligence and Lecture Notes in Bioinformatics)},
  volume = {13692},
  pages = {333-350},
  year = {2022},
  issn = {16113349, 03029743},
  isbn = {3031198239, 3031198247, 9783031198236, 9783031198243},
  publisher = {Springer Science and Business Media Deutschland GmbH},
  doi = {10.1007/978-3-031-19824-3_20}
}

@article{yu2024a_mip_splatting,
  author = {Yu, Zehao and Chen, Anpei and Huang, Binbin and Sattler, Torsten and Geiger, Andreas},
  title = {Mip-Splatting: Alias-Free 3D Gaussian Splatting},
  language = {eng},
  format = {article},
  journal = {Proceedings of the Ieee Computer Society Conference on Computer Vision and Pattern Recognition},
  pages = {19447-19456},
  year = {2024},
  issn = {2332564x, 10636919, 25757075},
  isbn = {9798350353006},
  publisher = {IEEE Computer Society},
  doi = {10.1109/CVPR52733.2024.01839}
}

@misc{Blender,
  title = {Blender - a 3D modelling and rendering package},
  organization = {Blender Foundation},
  address = {Blender Institute, Amsterdam},
  year = {2026},
  howpublished = {\url{https://www.blender.org/}},
}

@inproceedings{mallis2025cad,
  title={Cad-assistant: Tool-augmented vllms as generic cad task solvers},
  author={Mallis, Dimitrios and Karadeniz, Ahmet Serda and Cavada, Sebastian and Rukhovich, Danila and Foteinopoulou, Niki and Cherenkova, Kseniya and Kacem, Anis and Aouada, Djamila},
  booktitle={2025 IEEE/CVF International Conference on Computer Vision (ICCV)},
  pages={7284--7294},
  year={2025},
  organization={IEEE}
}

@inproceedings{wang2025vggt,
  title={VGGT: Visual Geometry Grounded Transformer},
  author={Wang, Jianyuan and Chen, Minghao and Karaev, Nikita and Vedaldi, Andrea and Rupprecht, Christian and Novotny, David},
  booktitle={Proceedings of the IEEE/CVF Conference on Computer Vision and Pattern Recognition},
  year={2025}
}

@software{harbor,
author = {{Harbor Framework Team}},
title = {{Harbor: A framework for evaluating and optimizing agents and models in container environments}},
year = {2026},
doi = {10.5281/zenodo.20953922},
howpublished = {\url{https://doi.org/10.5281/zenodo.20953922}}
}

@article{TwinTex23,

title={TwinTex: Geometry-aware Texture Generation for Abstracted 3D Architectural Models},

author={Weidan Xiong and Hongqian Zhang and Botao Peng and Ziyu Hu and Yongli Wu and Jianwei Guo and Hui Huang},

journal={ACM TOG (SIGGRAPH Asia)},

volume={42},

number={6},

pages={227:1--227:14},

year={2023},

}

@article{wang2021a,
  author = {Wang, Peng and Liu, Lingjie and Liu, Yuan and Theobalt, Christian and Komura, Taku and Wang, Wenping},
  title = {{NeuS:} Learning Neural Implicit Surfaces by Volume Rendering for Multi-view Reconstruction},
  language = {eng},
  format = {article},
  journal = {Advances in Neural Information Processing Systems},
  volume = {32},
  pages = {27171-27183},
  year = {2021},
  issn = {10495258},
  isbn = {1713845393, 9781713845393},
  publisher = {Neural information processing systems foundation}
}

@article{xiang2025trellis2,
    title={
Native and Compact Structured Latents for 3D Generation},
    author={
Xiang, Jianfeng and Chen, Xiaoxue and Xu, Sicheng and Wang, Ruicheng and Lv, Zelong and Deng, Yu and Zhu, Hongyuan and Dong, Yue and Zhao, Hao and Yuan, Nicholas Jing and Yang, Jiaolong},
    journal={
Tech report},
    year={
2025}
}

@article{kazhdan2006a,
  author = {Kazhdan, Michael and Bolitho, Matthew and Hoppe, Hugues},
  title = {Poisson surface reconstruction},
  language = {eng},
  format = {book},
  journal = {Proceedings of the Fourth Eurographics Symposium on Geometry Processing},
  pages = {61-70},
  year = {2006},
  doi = {10.5555/1281957.1281965}
}

@misc{european2024addressing,
  title={Addressing the environmental and climate footprint of buildings},
  author={{European Environment Agency}},
  year={2024},
  publisher={European Environment Agency, Copenhagen, Denmark},
  howpublished = {\url{https://build-up.ec.europa.eu/system/files/2024-10/TH-01-24-001-EN-N_Addressing-env-impact-buildings_FINAL.pdf}},
}

@article{finamore2026circularity,
  title={Circularity in construction: a multicriteria review and synthesis of existing frameworks and policy instruments},
  author={Finamore, Margherita and Oltean-Dumbrava, Crina},
  journal={Circular Economy and Sustainability},
  volume={6},
  number={2},
  pages={95},
  year={2026},
  publisher={Springer}
}

@article{groueix2018a,
  author = {Groueix, Thibault and Fisher, Matthew and Kim, Vladimir G. and Russell, Bryan C. and Aubry, Mathieu},
  title = {A Papier-Mache Approach to Learning 3D Surface Generation},
  language = {eng},
  format = {article},
  journal = {Proceedings of the Ieee Computer Society Conference on Computer Vision and Pattern Recognition},
  pages = {216-224},
  year = {2018},
  issn = {2332564x, 10636919, 25757075},
  isbn = {1538664208, 1538664216, 9781538664209, 9781538664216},
  publisher = {IEEE Computer Society},
  doi = {10.1109/CVPR.2018.00030}
}

@article{rakotosaona2021a,
  author = {Rakotosaona, Marie Julie and Guerrero, Paul and Aigerman, Noam and Mitra, Niloy and Ovsjanikov, Maks},
  title = {Learning Delaunay Surface Elements for Mesh Reconstruction},
  language = {eng},
  format = {article},
  journal = {Proceedings of the Ieee Computer Society Conference on Computer Vision and Pattern Recognition},
  pages = {22-31},
  year = {2021},
  issn = {25757075, 2332564x, 10636919},
  isbn = {1665445092, 9781665445092},
  publisher = {IEEE Computer Society},
  doi = {10.1109/cvpr46437.2021.00009, 10.1109/CVPR46437.2021.00009}
}

@misc{guertin2026drone,
  author       = {Matthew Guertin},
  title        = {Drone Building Scans},
  year         = {2026},
  howpublished = {\url{https://github.com/Matt1Up/drone-building-scans-dataset}},
  note         = {Dataset licensed under CC BY 4.0}
}
}


\end{document}